\documentclass[journal]{IEEEtran}
\ifCLASSINFOpdf
\else
\fi
\usepackage{times}
\usepackage{soul}
\usepackage{url}
\usepackage[hidelinks]{hyperref}
\usepackage[utf8]{inputenc}
\usepackage[small]{caption}
\usepackage{graphicx}
\usepackage{xcolor}
\usepackage{epsfig}

\usepackage{amsmath}
\usepackage{amssymb}
\usepackage{mathrsfs}
\usepackage{txfonts}
\usepackage{bm}
\usepackage{multirow}
\usepackage{tabularx}
\usepackage{caption,subcaption}
\usepackage{booktabs}
\usepackage{algorithm}
\usepackage{algorithmic}
\usepackage{float}
\usepackage{array}

\usepackage{cite}
\usepackage{makecell}

\usepackage{xcolor}
\usepackage{graphicx} %use graph format
\usepackage{epstopdf}

\begin{document}
\title{BP-Triplet Net for Unsupervised Domain Adaptation: A Bayesian Perspective}

\author{Shanshan Wang,
        Lei Zhang,~\IEEEmembership{Senior Member,~IEEE,~}
        Pichao Wang% <-this % stops a space
\IEEEcompsocitemizethanks{\IEEEcompsocthanksitem
%This work was supported by the National Science Fund of China under Grants (61771079, 61401048) and the Fundamental Research Funds for the Central Universities (No. 106112017CDJQJ168819).

S. Wang is with the Institutes of Physical Science and Information Technology, Anhui University, Anhui 230601, China.
\ (E-mail: wang.shanshan@ahu.edu.cn).
 \IEEEcompsocthanksitem L. Zhang is with the College of Microelectronics and Communication Engineering, Chongqing University,  Chongqing 400044, China.% <-this % stops an unwanted space
 \ (E-mail: leizhang@cqu.edu.cn).
 \IEEEcompsocthanksitem P. Wang is with Alibaba Group, Bellevue, WA, 98004, USA.% <-this % stops an unwanted space
 \ (E-mail: pichao.wang@alibaba-inc.com).
}}% <-this % stops a space


\markboth{journal submission}%
{Shell \MakeLowercase{\textit{et al.}}: Bare Demo of IEEEtran.cls for IEEE Journals}



\maketitle

\begin{abstract}
Triplet loss, one of the deep metric learning (DML) methods, is to learn the embeddings where examples from the same class are closer than examples from different classes. Motivated by DML, we propose an effective \textbf{BP-Triplet Loss} for unsupervised domain adaption (UDA) from the perspective of Bayesian learning and we name the model as \textbf{BP-Triplet Net}. In previous metric learning based methods for UDA, sample pairs across domains are treated equally, which is not appropriate due to the domain bias. In our work, considering the different importance of pair-wise samples for both feature learning and domain alignment, we deduce our BP-Triplet loss for effective UDA from the perspective of Bayesian learning. Our BP-Triplet loss adjusts the weights of pair-wise samples in intra-domain and inter-domain. Especially, it can self attend to the hard pairs~(including hard positive pair and hard negative pair). Together with the commonly used adversarial loss for domain alignment, the quality of target pseudo labels is progressively improved. Our method achieved low joint error of the ideal source and target hypothesis. The expected target error can then be upper bounded following Ben-David's theorem.
Comprehensive evaluations on five benchmark datasets,  handwritten digits, Office31, ImageCLEF-DA, Office-Home and VisDA-2017 demonstrate the effectiveness of the proposed approach for UDA.
\end{abstract}

\begin{IEEEkeywords}
Cross Domain Class Alignment, Unsupervised Domain Adaptation, Metric Learning, Bayesian Perspective.
\end{IEEEkeywords}

% \fi
\IEEEpeerreviewmaketitle

\section{Introduction}
\begin{figure}
\begin{center}
 \includegraphics[width=0.9\linewidth]{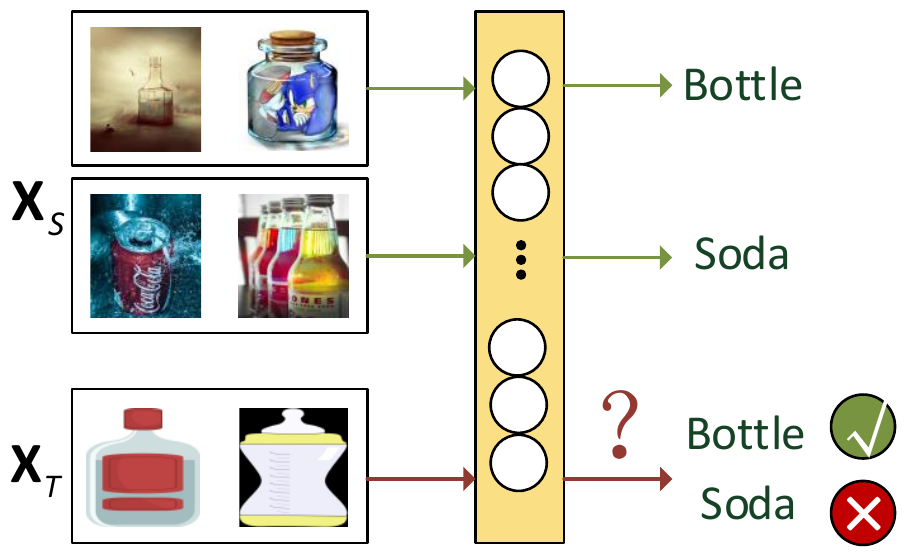}
\end{center}
  \setlength{\abovecaptionskip}{0pt}
   \caption{Illustration of domain shift problem.}
\label{sketch}
\end{figure}

\IEEEPARstart{D}{eep} networks have established the state-of-the-arts for diverse visual applications significantly, such as person re-identification~\cite{yang2018person}, video retrieval~\cite{dong2021dual,yang2020tree,yang2022video,yang2021deconfounded}, visual classification~\cite{tan2021selective,meng2019learning}. However, it is still a big challenge for generalizing the learned knowledge to new domains.
The classifier and the convolutional layers trained on source images may not generalize well on target samples.
Domain Adaptation~(DA)~\cite{pan2010survey,He2015Deep,zhang2020adversarial,zhang2019optimal,zhang2019image} is an effective way to address the domain shift problem. It aims to learn from a sufficiently labeled domain~(known as source domain) and to solve the same problem in the other related but different domain~(target domain), where a few even no labeled samples are available~\cite{Hoffman2014Asymmetric}.
%recognize the unlabeled target domain leveraging a sufficiently labeled, related but different source domain~\cite{tzeng2015simultaneous,oquab2014learning,Hoffman2014Asymmetric}.
In this paper, we address the challenging but practical topic, \textit{i.e.},  unsupervised domain adaptation
(UDA)~\cite{Krizhevsky2012ImageNet}. Taken Figure~\ref{sketch} as an example, this task aims to recognize the unlabeled target domain data leveraging a sufficiently labeled, related but different source domain~\cite{tzeng2015simultaneous,oquab2014learning}. The key issue of DA problem is to reduce distribution difference between two domains, such that the learned classifier from source domain can well classify target domain samples.

Recently, domain adaptation has witnessed significant progress in algorithms~\cite{xie2015transfer,SimonKornblith2018deep,csurka2017domain,ajakan2014domain}. Domain-invariant or domain-confused feature representation is usually thought to have been learned.
%,bousmalis2016domain,ganin2017domainMany works focus on distribution alignment between domains by mapping both domains into a new subspace~\cite{He2015Deep,Krizhevsky2012ImageNet,tzeng2015simultaneous,oquab2014learning,xie2015transfer,SimonKornblith2018deep}.
Generally, maximum mean discrepancy (MMD)~\cite{gretton2012kernel}, as a non-parametric metric method, is commonly used to measure the dissimilarity of distributions. After MMD guided distribution alignment, adversarial learning~\cite{ganin2017domain,bousmalis2016domain} as another mainstream has been successfully brought into DA approaches to reduce distribution discrepancy. Unlike many previous MMD-based DAs, domain-adversarial neural networks focus on combining DA and deep feature learning within a unified training paradigm.
The goal of adversarial domain adaptation is to confuse the features between domains, so that domain-invariant representations are ultimately obtained.
In this way, the obtained feed-forward network can be applicable to the target domain without being hindered by the shift between two domains.
%In unsupervised domain adaptation, we have only unlabeled training samples in the target domain.
In this paper, to learn domain invariant representations, we adopt the structure of adversarial network. We add one domain classifier at the last feature extraction block and learn domain invariant features by minimax optimization between the domain classifier and the feature representation network. The gradient reversal layer (GRL) can be utilized during the backpropagation process.

However, only adopting adversarial learning does not really guarantee safe domain alignment. On one hand, the alignment of category space between domains is ignored in alleviating domain shift. Target samples that are close to the decision boundary are easily to be misclassified by the source classifier~\cite{wen2016discriminative}. On the other hand, in practical domain adaptation problems, the data distributions of the source and target domain usually embody complex multimode structures. Thus, previous domain adaptation methods that aim to reduce the domain distributions without exploiting the class-wise structures may lead to negative transfer.
%For one thing, when data distributions embody complex multimodal structures, adversarial adaptation methods may fail to capture such multimodal structures for a discriminative alignment of distributions without mode mismatch. For another,

\begin{figure}
\begin{center}
 \includegraphics[width=1.0\linewidth]{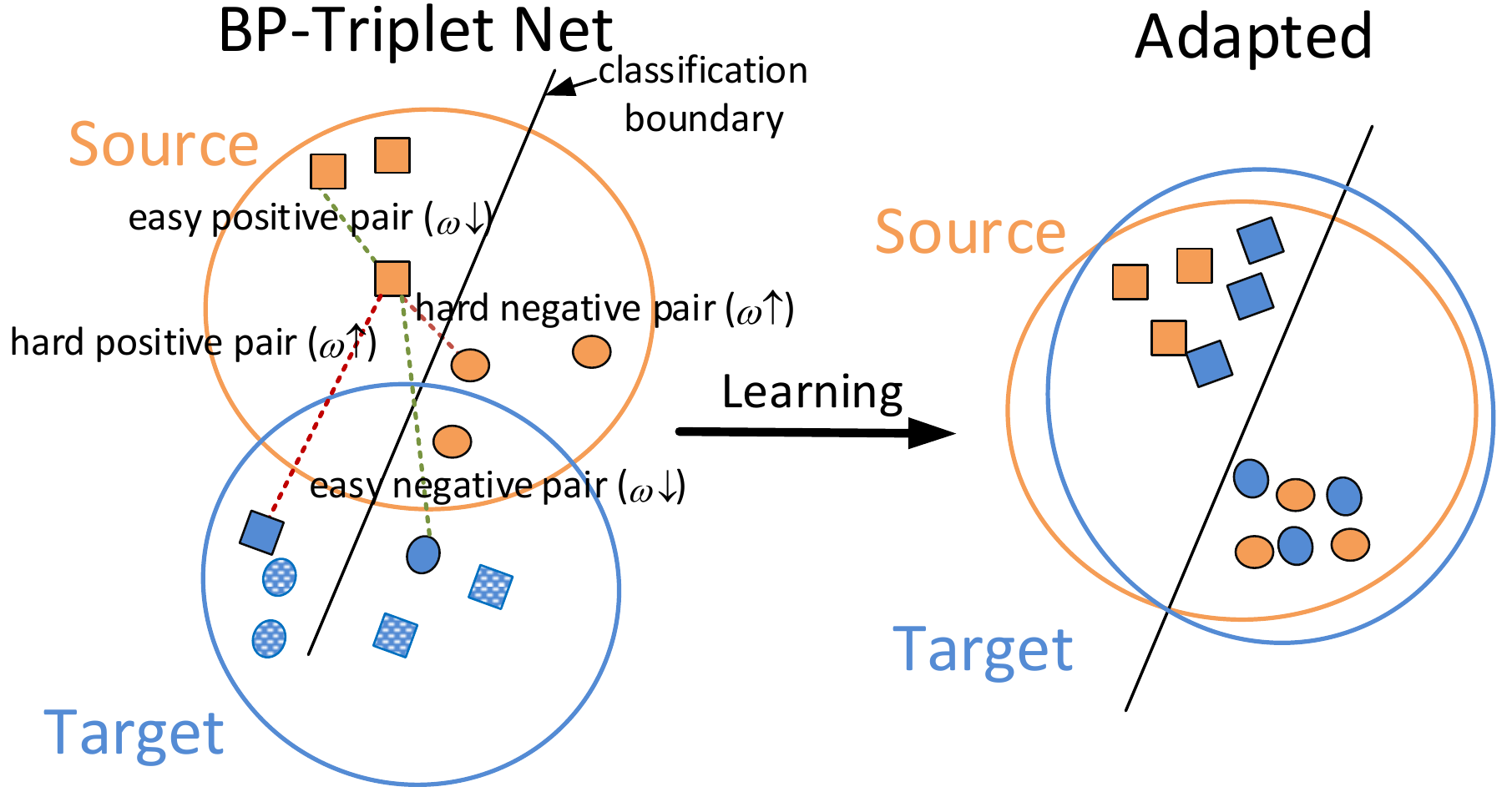}
\end{center}
  \setlength{\abovecaptionskip}{0pt}
   \caption{Motivation of our method. The domain shift explicitly exists before domain adaptation. To avoid the misalignment, by minimizing intra-class compactness and maximizing inter-class separation, domain adaptation is achieved. Specifically, we assign different weights to different sample pairs based on the assumption that hard pairs can improve cross-domain generalization performance. We let $\omega$ express weight, then down-weight~($\omega\downarrow$) easy pairs and up-weight~($\omega\uparrow$) hard pairs according to pair-wise importance. Our method performs class-aware alignment through similarity learning across domains. Circle and block represent two different classes.  Different colors mean different domains. Shadow denotes the misclassified samples. The same (different) labeled samples which have a large (small) distance are named as hard pairs, and the same (different) labeled samples that have a small (large) distance are named as easy pairs.}
\label{figcomparision}
\end{figure}

To deal with this challenge, we adopt the general idea of class-wise relations by leveraging similarity learning  to promote domain adaptation.
As shown in Figure~\ref{figcomparision},
%this work devote to mitigating the domain shift by domain alignment.
%all the domain alignment approaches can only reduce but not remove the domain discrepancy.However,
considering domain bias, target samples near the edge of clusters, or far from their corresponding class centers are most likely to be misclassified. The methods which aim at reducing domain shift by feature alignment through a discrepancy metric but ignore class-wise discrimination are not enough.
To alleviate this issue, a practical way is to enforce the samples with better intra-class compactness and inter-class separation. In computer vision, metric learning algorithms have been proposed to learn the similarity, which nicely meets our need to force feature matching.
In this way, the number of misclassified samples will be greatly reduced.
The motivation of the metric loss is two-fold. On one hand, samples with the same labels should be pulled together in the embedding space. On the other hand, samples with different labels should be far away from each other.
Thus, in addition to learning the domain invariant representations, we further adopt and revisit the metric learning method for UDA task. In practice, triplet loss~\cite{Schroff2015FaceNet,wang2020self} that as an effective metric learning loss can be applied in our model and it constructs triplet sample pairs guided by labels.

In this paper, the same (different) labeled samples which have  large (small) distance are named as hard pairs, and the same (different) labeled samples that have  small (large) distance are named as easy pairs.
In cross-domain problems, although the sample pairs come from the same class, they may also come from different domains.~\textit{e.g.}, the easy positive pairs tend to come from the same domain, and the hard positive pairs come from different domains. Noteworthily, the hard positive pairs are key issues in cross domain problem.

Considering that the pair-wise importance imbalance is noteworthy in cross-domain problems, and motivated by focal loss~\cite{Lin2017Focal} and Maximum A Posteriori estimation~(MAP)~\cite{yang2011robust}, we propose a BP-triplet loss to deal with these problems.
We propose to address the pair-wise problem by reshaping the standard sample pairs such that it down-weights the loss assigned to easy pairs and up-weights hard pairs. This is because easy positive examples tend to have small distance or easy negative examples are likely to have large distance. \textbf{ Specifically, we model our loss function as the solution of MAP and name it as BP-triplet loss. It self attends to hard pairs rather than imposing equal importance for all sample pairs. Consequently, the class alignment across domains is promoted by strengthening hard pairs and simultaneously weakening easy pairs.}

Additionally, to learn domain invariant representations, we adopt the structure of adversarial network. We add one domain classifier at the last feature extraction block, and learn domain invariant features by minimax optimization between the domain classifier and the feature representation network.
Thus, our model can learn robust representations which are not only domain invariant for domain alignment but also class discriminative for class alignment. This will have a more confident guarantee that the joint error of the ideal source and target hypothesis is low. The DA becomes possible as presented in Ben-David's theorem.

The main contributions and novelties of this paper are summarized as follows.
\begin{itemize}
\item We design a BP-triplet net to learn robust representations. The representations are not only domain invariant but also class discriminative for semantic alignment by leveraging metric learning. Further more, different from previous metric learning methods, we use the constraint of metric loss not only to align class relations but also to align the domain features by up-weighting the cross domain hard positive pairs.
\item In order to deal with pair-wise importance imbalance and enhance domain confusion during class-wise feature alignment, we propose a BP-triplet loss to reduce the weight of easy pairs while increasing the hard pair weight. We force those poorly aligned samples to be clustered. Noteworthily, different from the integration of various losses, our loss function is deduced from the perspective of Bayesian learning  and MAP, which is a novel technical contribution.
\item  Together with the commonly used adversarial loss, the proposed model achieves promising results on four datasets, which clearly demonstrates the effectiveness of the proposed method for UDA.

%  To facilitate the training of class level alignment, we employ a class-aware sampling strategy from both source and target domains and sample data within a randomly sampled class subset.
\end{itemize}

\section{Related Works}

\subsection{Metric Learning}
%\textbf{Metric Learning.}
In recent years, a variety of metric learning algorithms~\cite{yang2017person,xun2016empirical,liu2020deep} have been proposed in the literature. Specifically, contrastive loss~\cite{Hadsell2006Dimensionality} and triplet loss~\cite{Schroff2015FaceNet} have been intensively used in UDA. \textit{e.g.}, SimNet~\cite{pinheiro2018unsupervised} proposes to classify an image by computing its similarity to prototype representations of each category. Deng~\textit{et al.} propose a SCA method~\cite{Deng2018Domain} by jointly learning domain and class alignment through MMD and metric learning. In~\cite{chen2018joint}, JDDA is proposed to achieve domain alignment and discriminative feature learning jointly. Even though these works leverage the metric learning in their methods, there is no supervision for domain alignment in metric learning. Different from these methods, we leverage the constraint of metric learning not only to align class relations but also to align the domain features.

\subsection{Deep Domain Adaptation}
%\textbf{Deep Domain Adaptation.}
As deep representations can only reduce, but not remove the cross-domain discrepancy. Recent research on deep domain adaptation further embeds domain-adaptation modules in deep networks to boost transfer performance.

In UDA, Training CNN can be conducted through various strategies. Matching distributions of the extracted features in CNN is considered to be effective for an accurate adaptation. These works mainly pay attention to the statistics alignment.
Tzeng et al.~\cite{tzeng2015simultaneous} propose a DDC method and achieve successful knowledge transfer between domains and tasks. Long et al.~\cite{long2015learning} propose a deep adaptation network (DAN) by imposing MMD loss on the high-level features across domains. Additionally, Long et al.~\cite{long2017deep} also propose joint maximum mean discrepancy (JMMD) to measure the relationship of joint distribution. These works focus more on feature alignment. Another impressive work is DeepCORAL~\cite{sun2016deep}, which extends CORAL to deep architectures and aligns the high-order covariance statistics between the source and target features.
Zellinger et al. propose a CMD method~\cite{zellinger2017central} to match the higher order central moments of probability distributions via order-wise moment differences.

Recently, Generative Adversarial Nets (GANs)~\cite{Goodfellow2014Generative} inspired adversarial domain adaptation methods have been preliminarily studied. In~\cite{ganin2017domain}, DANN method proposes domain adversarial learning, in which a gradient reversal layer is designed for confusing features from two domains by minimax optimization between the network and a domain classifier. This method can be regarded as the baseline of adversarial learning methods. Tzeng~\textit{etal.} proposed an ADDA method~\cite{Tzeng2017Adversarial} which combines discriminative modeling, untied weight sharing and a GAN loss.
The work has shown the potential of adversarial learning in domain adaptation.
In~\cite{Hoffman2017CyCADA}, Hoffman~\textit{et al.} propose a CYCADA method which adapts representations at both the pixel-level and feature-level, enforcing cycle-consistency by leveraging a task loss.
In~\cite{saito2017maximum}, Saito et al. propose a new approach called MCD that uses two different classifiers to align those easily misclassified target samples through adversarial learning in CNN.
Zhang~\textit{et al.}~\cite{zhang2018collaborative} propose a CAN by imposing domain classifiers on multiple CNN blocks to learn domain invariant representations through adversarial learning.
%learn domain informative representations from lower blocks through collaborative learning and
Long~\textit{et al.}~\cite{Long2017Conditional} also present a conditional adversarial domain adaptation (CDAN) that conditions the discriminative information conveyed into the predictions of classifier.

However, these methods are based on the theory that the predicted error is bounded by the distribution divergence. They do not consider the relationship between target samples and decision boundaries. To tackle these problems, we propose a strategy leveraging the target samples. Additionally, different from the previous metric learning, our BP-triplet loss not only aligns class relations but also to align the domain features.

 \begin{figure}
\begin{center}
 \includegraphics[width=1\linewidth]{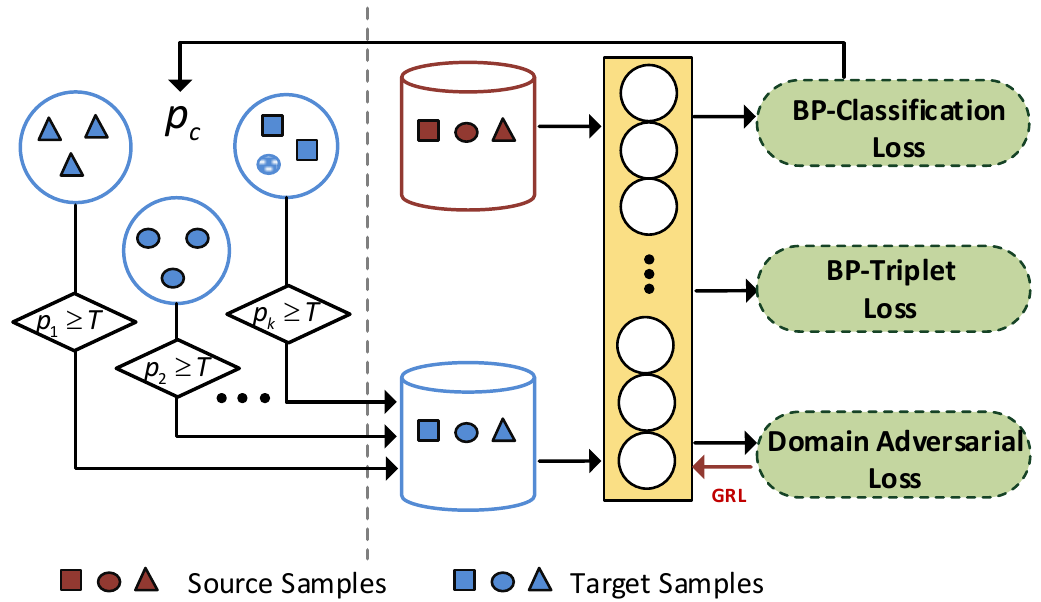}
\end{center}
  \setlength{\abovecaptionskip}{0pt}
   \caption{The framework of our method which includes three parts: 1) a BP-triplet loss is deduced from Bayesian formulation for category and domain alignment by a specially designed triplet pairing strategy; 2) the prior induced cross-entropy loss is used for source classifier training; 3) adversarial loss with GRL is deployed for domain feature alignment.}
\label{fig1}
\end{figure}

\section{The Proposed Method}
\subsection{Notation}
Our goal is to predict the target label $\hat{y}^t = \arg \max {G_y}(f(x^t))$ and an overview of our method is depicted in Figure \ref{fig1}. We suppose $\mathcal{D}_s=\{(x_i^s,y_i^s)\}_{i=1}^{n_{s}}$ and $\mathcal{D}_t=\{x_j^t\}_{j=1}^{n_{t}}$ to be the labeled source data and unlabeled target data, drawn from different distributions respectively.
Our method aims to reduce the domain gap through minimizing the source risk $\epsilon_s(G_y)=\mathbb{E}_{(\bm{x}^s,\bm{y}^s)\sim P}[G_y (f(\bm{x}^s))\neq \bm{y}^s]$ confused by both domain and class-wise alignment, such that the target risk $\epsilon_t(G_y)=\mathbb{E}_{(\bm{x}^t,\bm{y}^t)\sim Q}[G_y (f(\bm{x}^t))\neq \bm{y}^t]$ can be minimized, where $G_y(\cdot)$ represents the softmax output and $ f(\cdot) $ refers to the feature representation.

\subsection{Model formulation of BP-triplet Net}
Our model is composed of three components: 1) BP-triplet loss for feature similarity learning across domains, 2) classification loss for classifier training, and 3) adversarial loss for domain distribution alignment. Noteworthily, the first two items are deduced from a Bayesian learning perspective.

%\subsection{Bayesian Formulation.}
\textbf{Bayesian Formulation.}
%In this section, we give the construction of our model based on Bayesian decision theory.
Suppose the hybrid domain $\mathcal{D}=\mathcal{D}_s\cup\mathcal{D}_t$, and $\hat{y}^t$ in $\mathcal{D}$ is the pseudo target label predicted by the source classifier. Given a triplet ($x_i, x_j, x_k)\in \mathcal{D}$, let $s_{i,j}$ represent the pair-wise similarity between $x_i$ and $x_j$. $s_{i,j}= 1$ means they have the same label; otherwise, $s_{i,j}=0$ means they have different labels.

Without loss of generality, let ${p}(f_i ,f_j ,f_k|s_{i,j} ,s_{i,k} )$ be the posterior probability of feature representation $f_i$, $f_j$, $f_k$ for triplet sample set $x_i, x_j, x_k$.
With the assumption of conditional independence of each pair and Bayesian formula, the joint posterior probability density function of the pair-wise training set can be generally represented as
\setlength\abovedisplayskip{2pt}
\setlength\belowdisplayskip{2pt}
 \begin{equation}
\begin{split}
 &\prod\limits_{i,j,k \in \mathcal{D}} {p(f_i ,f_j ,f_k|s_{i,j} ,s_{i,k} )}\Leftrightarrow \\
 &\prod\limits_{i,j,k \in \mathcal{D}} {p(s_{i,j} ,s_{i,k}|f_i ,f_j ,f_k )} p(f_i )p(f_j )p(f_k ),\\
\end{split}
\label{eequa1}
\end{equation}
where ${p(s_{i,j} ,s_{i,k}|f_i ,f_j ,f_k )}$ is the likelihood probability and $p(f)$ represents the prior probability of feature representation for each sample.
%where ${p(s_{i,j} ,s_{i,k} /f_i ,f_j ,f_k )}$ is the likelihood function and we design a function as follows to represent it:
%\begin{equation}
%\begin{split}
%&p(s_{i,j} ,s_{i,k} /f_i ,f_j ,f_k ) = \\
%&(p_1^{s_{i,j} a} p_2^{(1 - s_{i,j} )a} )^a (p_3^{s_{i,j} (1 - a)} p_4^{(1 - s_{i,j} )(1 - a)} )^{(1 - a)},
%\end{split}
%\label{eequa8}
%\end{equation}
%where $a$ is a XOR expression by $a = s_{i,j}\oplus s_{i,k}$.

Our aim is to find the optimal model parameters $\theta_f$ of feature representation $f$, which is the solution of maximum a posteriori estimation~(MAP) of Eq.~(\ref{eequa1}) from the Bayesian perspective.
%\begin{equation}
%\begin{split}
%&\max P(F/S) = \max \prod\limits_{i,j,k \in P} {p(f_i ,f_j ,f_k /s_{i,j} ,s_{i,k} )} \\
%&= \max \prod\limits_{i,j,k \in P} {p(s_{i,j} ,s_{i,k} /f_i ,f_j ,f_k )} p(f_i )p(f_j )p(f_k ).
%\end{split}
%\label{eequa10}
%\end{equation}
More formally, we propose to add a modulating factor $\omega$ to the likelihood probability, with tunable focusing
parameter $\gamma\ge0$, formulated as
\begin{equation}
\begin{split}
&\omega  = (1 - p(s_{i,j} ,s_{i,k}|f_i ,f_j ,f_k ))^\gamma.\\
\end{split}
\label{eequa2}
\end{equation}

Intuitively, the modulating factor reduces the loss contribution from easy pairs and penalizes more on those hard pairs. For the convenience, the same (different) labeled samples which have a large (small) distance are named as hard pairs, and the same (different) labeled samples that have a small (large) distance are named as easy pairs.

Therefore, the MAP of the joint posterior probability in Eq. (\ref{eequa1}) with the modulating factor $\omega$ in Eq. (\ref{eequa2}) of the likelihood probability can be formulated as:
\begin{equation}
\begin{split}
&\mathop {\max}\limits_{\theta _f} \prod\limits_{i,j,k \in \mathcal{D}} {p(s_{i,j} ,s_{i,k}|f_i ,f_j ,f_k )}^\omega p(f_i )p(f_j )p(f_k ),\\
&\propto \mathop {\min}\limits_{\theta _f}~~\sum\limits_{i,j,k \in \mathcal{D}}-{\omega \log p(s_{i,j} ,s_{i,k}|f_i ,f_j ,f_k )}  \\
&~~-\sum\limits_{i \in \mathcal{D}}{\log p(f_i )}-\sum\limits_{j \in \mathcal{D}}{\log p(f_j )}-\sum\limits_{k \in \mathcal{D}}{\log p(f_k)}.
\end{split}
\label{eequa3}
\end{equation}

%Maximizing the posteriori probability is equivalent to minimize its negative log-likelihood function,
%\setlength\abovedisplayskip{2pt}
%\setlength\belowdisplayskip{2pt}
%\begin{equation}
%\begin{split}
%&\mathop {\min}\limits_{\theta _f}~~\sum\limits_{i,j,k \in \mathcal{D}}-{\omega \log p(s_{i,j} ,s_{i,k}|f_i ,f_j ,f_k )}  \\
%&~~-\sum\limits_{i \in \mathcal{D}}{\log p(f_i )}-\sum\limits_{j \in \mathcal{D}}{\log p(f_j )}-\sum\limits_{k \in \mathcal{D}}{\log p(f_k)}.
%\end{split}
%\label{eequa12}
%\end{equation}

%\subsection{Focal-triplet Loss.}
\textbf{BP-Triplet Loss.}
Now we focus on the first item of Eq.~(\ref{eequa3}).
According to the sample pair similarity, we explore the likelihood probability function of ${p(s_{i,j} ,s_{i,k}|f_i ,f_j ,f_k )}$ by supposing it to be exponential distribution
due to the good convergence. Then the likelihood probability density function can be represented as:
%instead of sigmoid function, due to the better convergence of exponential function. Then the likelihood probability density function can be represented as:
 \setlength\abovedisplayskip{2pt}
 \setlength\belowdisplayskip{2pt}
 \begin{equation}
\begin{split}
&p(s_{i,j} ,s_{i,k}|f_i ,f_j ,f_k ) = \\
&\left\{ {\begin{array}{*{20}c}
   {e^{-\alpha(d_{i,j}-d_{i,k}+m)},~~\mathrm{if}~~s_{i,j}=1,s_{i,k}=0}  \\
   {e^{-\alpha(-d_{i,j}+d_{i,k}+m)},\mathrm{if}~~s_{i,j}=0,s_{i,k}=1}  \\
   {e^{-\alpha(d_{i,j}+d_{i,k})},~~~~~~~~\mathrm{if}~~s_{i,j}=1,s_{i,k}=1}  \\
   {e^{-\alpha(-d_{i,j}-d_{i,k})},~~~~~~\mathrm{if}~~s_{i,j}=0,s_{i,k}=0},  \\
\end{array}} \right.
\end{split}
\label{eequa4}
\end{equation}
where $\alpha$ is a hyper-parameter of the exponential function. $d_{i,j}$ represents the distance between the feature representations $f_i$ and $f_j$. In this paper, the Euclidean distance is considered, \textit{i.e.}, $d_{i,j}  = \left\| {f_i - f_j } \right\|^2 $. Similarly, $d_{i,k}  = \left\| {f_i - f_k } \right\|^2 $. $m$ is a margin which is enforced between positive and negative pairs and it can be preset.

For simplification, the general formulation of the likelihood probability in Eq.~(\ref{eequa4}) can be further written as:
\setlength\abovedisplayskip{2pt}
\setlength\belowdisplayskip{2pt}
\begin{equation}
\begin{split}
&p(s_{i,j} ,s_{i,k}|f_i ,f_j ,f_k )=\\
&e^{\alpha (( - 1)^{s_{i,j} } d_{i,j}  + ( - 1)^{s_{i,k} } d_{i,k}  - \beta\cdot m)},
\end{split}
\label{eequa5}
\end{equation}
where $\beta = s_{i,j}\oplus s_{i,k}$, and $\oplus$ is the XOR operator.

By substituting Eq.~(\ref{eequa5}) into the first item of Eq.~(\ref{eequa3}), the weighted log-likelihood probability can be written as
\setlength\abovedisplayskip{2pt}
\setlength\belowdisplayskip{2pt}
\begin{equation}%\small%\footnotesize
\begin{split}
&\sum\limits_{i,j,k \in \mathcal{D}}- {\omega \log p(s_{i,j} ,s_{i,k}|f_i ,f_j ,f_k )}= \\
% = &\max\sum\limits_{i,j,k \in P}{\omega \log (p_1^{s_{i,j} a} p_2^{(1-s_{i,j} )a} )^a (p_3^{s_{i,j} (1-a)} p_4^{(1-s_{i,j} )(1-a)})^{(1-a)}}\\
%\sum\limits_{i,j,k \in \mathcal{S}}{\omega \alpha m(1-a))}\\
&\sum\limits_{i,j,k \in \mathcal{D}}-{\omega \alpha\lbrack( - 1)^{s_{i,j} } d_{i,j}  + ( - 1)^{s_{i,k} } d_{i,k}  - \beta\cdot m\rbrack_+},\\
%&+\sum\limits_{i,j,k \in \mathcal{S}}{\omega \alpha m(1-a))}\\
\end{split}
\label{eequa6}
\end{equation}
where $\lbrack x\rbrack_+$ denotes the operator of $\max(x,0)$, which is manually imposed for improving the convergence.

In model optimization, we consider the case that positive pairs and negative pairs both exist. Also, in constructing triplet pairs, we can set $x_i$ as the anchor, $x_j$ is similar to the anchor, and $x_k$ is dissimilar to the anchor. Therefore, the condition in Eq.~(\ref{eequa6}) can be relaxed as $s_{i,j}= 1, s_{i,k}= 0$. For the convenience, by substituting Eq.~(\ref{eequa2}) into Eq.~(\ref{eequa6}), we can rewrite the simplified focal-triplet loss as:
\setlength\abovedisplayskip{2pt}
\setlength\belowdisplayskip{2pt}
\begin{equation}
\begin{split}
&{{\mathcal{L}}_{{\rm{BP-tri}}}}(\theta _f)=  \\
&\sum\limits_{i,j,k \in \mathcal{D}} {\alpha(1 - e^{-\alpha(d_{i,j}- d_{i,k}  + m)})^\gamma \lbrack d_{i,j}- d_{i,k}+ m\rbrack_+}.
\end{split}
\label{eequa7}
\end{equation}
\setlength\belowdisplayskip{2pt}

Obviously, it is a variant of standard triplet loss, by down-weighting easy pairs and up-weighting hard pairs to impose different importance for different sample pairs.
 \begin{figure}
\begin{center}
 \includegraphics[width=1\linewidth]{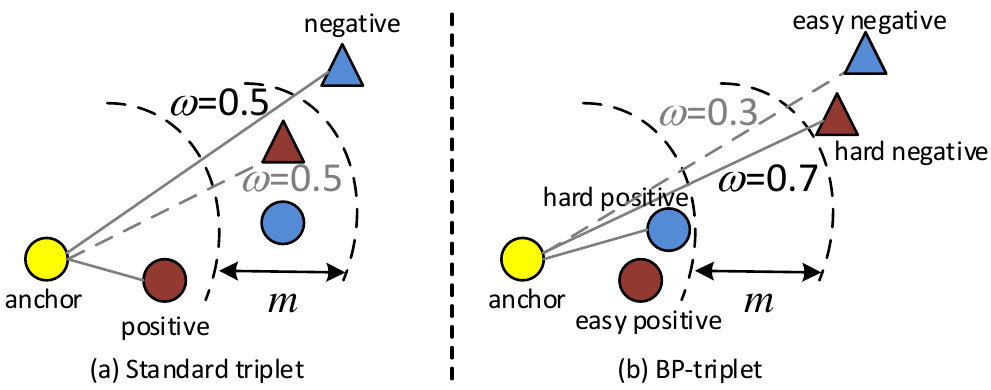}
\end{center}
  \setlength{\abovecaptionskip}{0pt}
   \caption{Illustration of the proposed BP-triplet. Given an anchor and its ranked list, the anchor is compared with one negative example and one positive example in standard triplet. In this case, some non-trivial structured information is ignored or merely extracted because the sample pairs have the same weights. To address it, our BP-triplet can down-weight easy pairs and up-weight hard pairs. \textit{e.g.}, BP-triplet enforces the anchor far away from the hard negative examples than the easy negative examples, and the distance to the cross-domain negatives can be maximized.}
\label{fig2}
\end{figure}

%\subsection{Classification Loss.}
\textbf{BP-Classification Loss.}
Then we turn to the last three items in Eq.~(\ref{eequa3}). Because $x_i, x_j, x_k$ are from domain $\mathcal{D}=\mathcal{D}_S\cup\mathcal{D}_T$ and there is $-\sum_{i\in\mathcal{D}}\log p(f_i)=-\sum_{j\in\mathcal{D}}\log p(f_j)=-\sum_{k\in\mathcal{D}}\log p(f_k)$. Therefore, we focus on one item $-\sum_{i\in\mathcal{D}}\log p(f_i)$ for convenience. For formulating the prior probability, we divide it into two parts, \textit{i.e.}, $-\sum_{i\in\mathcal{D}}\log p(f_i)=-\sum_{i\in\mathcal{D}_S}\log p(f_i)-\sum_{i\in\mathcal{D}_T}\log p(f_i)$. Intuitively, for keeping the consistency with the neural network training, we adopt the softmax function to compute the prior probability $p(f_i)$. For the sufficiently labeled source domain, the labels can be used to guide the computation of prior.
For the unlabeled target domain, the entropy minimization principle is employed to enhance discrimination of learned models for target data.
Therefore, the prior probability induced classification loss ${{\mathcal{L}}_{c}}$ can be formulated as
 \setlength\abovedisplayskip{2pt}
 \setlength\belowdisplayskip{2pt}
\begin{equation}
\begin{split}
&\mathcal{L}_c(\theta _f,\theta _y)=-\sum_{i\in\mathcal{D}}\log p(f_i)\\
&=-\sum_{i\in\mathcal{D}_S}y^s_i\log p(f_i)-\sum_{i\in\mathcal{D}_T}p(f_i)\log p(f_i),\\
%{\mathcal{L}}_c(\theta _f,\theta _y) = \frac{1}{{n_s+n_t}}&(\sum\limits_{{\bf{x}}_i \in \mathcal{D}_s } {{\mathcal{L}}_y } (G_y (f ({\bf{x}}_i )),{\bf{y}}_i ) +\\
%&\sum\limits_{{\bf{x}}_j \in \mathcal{D}_t } { - \log {G_y (f ({\bf{x}}_j )}})).
\end{split}
\label{eequa8}
\end{equation}
where $y^s_i$ represents the label of source sample $x_i$. Obviously, the classification loss is the general cross-entropy loss and conditional entropy loss in convolutional neural network.

%&\sum\limits_{{\bf{x}}_j \in \mathcal{D}_t } { - \log \sum\limits_{k = 1}^K {G_y (f ({\bf{x}}_j ))_k}}).

%Now we combine the two parts (Eq.~(\ref{eequa16}) and Eq.~(\ref{eequa14})) in Eq.~(\ref{eequa12}). Then, the similarity learning of Eq.~(\ref{eequa12}) can be expressed as:
%\setlength\abovedisplayskip{0pt}
%\setlength\belowdisplayskip{0pt}
%\begin{equation}%\small
%\begin{split}
%\mathop {\min}\limits_{\theta _f,\theta _y}~&{{\mathcal{L}}_{Sim}} ={{\mathcal{L}}_{{\rm{F-tri}}}}+{{\mathcal{L}}_{c}}\\
%%=&\sum\limits_{i \in \mathcal{D}_s} { - \bm{y}_i \log p(f_i )}+\sum\limits_{j \in \mathcal{D}_t } { - \log \sum\limits_{k = 1}^K {p_k (f_i )}}+\\
%%&\sum\limits_{i,j,k \in \mathcal{S}} {(1 - e^{-(d_{i,j}- d_{i,k}  + m)})^\gamma (d_{i,j}- d_{i,k}+ m)}.
%\end{split}
%\label{eequa15}
%\end{equation}

%\subsection{Domain Adversarial Loss.}
\textbf{Domain Adversarial Loss.}
%\subsection{Notation}
So far, we have not discussed the distribution alignment problem. In UDA setting,
%\subsection{Model Formulation}
%\textbf{Domain Adversarial Networks.}
%$ G_d(\cdot)$ means the domain discriminator.
domain adversarial networks~\cite{ganin2017domain,bousmalis2016domain,wang2020self} have been explored to minimize the domain discrepancy and simultaneously extract transferable features. The procedure is a two-player game: the first player is the domain discriminator $G_d$ trained to distinguish the source domain from the target domain, and the second player is the feature extractor $f(\cdot)$ (\textit{e.g.}, neural network) trained to confuse the domain discriminator. The minimax objective of adversarial loss is:
\setlength\abovedisplayskip{2pt}
\setlength\belowdisplayskip{2pt}
\begin{equation}\small
\begin{split}
\mathop {\min }\limits_{\theta_f }&\mathop {\max }\limits_{\theta_d } ~~{\mathcal{L}}_{adv}(\theta _f, \theta _d ) =\\
& -\frac{1}{n_s+n_t}\sum\limits_{{{x}}_i  \in (\mathcal{D}_s  \cup \mathcal{D}_t )} {{\mathcal{L}}_d } (G_d (f ({{x}}_i )),d_i ),\\
\end{split}
\label{eequa9}
\end{equation}
%where ${n=n_s+n_t}$.
where $\theta_f, \theta_d$ represent the parameters of feature network and domain discriminator, respectively. $d_i$ is the domain label of sample $x_i$.
%$\mathcal{L}_d$ is the cross-entropy loss of $G_d$.
%\setlength\abovedisplayskip{2pt}
%\setlength\belowdisplayskip{0pt}
%\begin{equation}%\small
%\begin{array}{l}
% \mathop {(\theta_f }\limits^\wedge) = \arg \mathop {\min }\limits_{\theta_f } {\mathcal{L}}_{adv}(\theta_f,\theta_d ) \\
% \mathop {(\theta_d }\limits^\wedge) = \arg \mathop {\max }\limits_{\theta_d } {\mathcal{L}}_{adv}(\theta_f,\theta_d ). \\
% \end{array}
%\label{eequa2}
%\end{equation}

%\subsection{Overall Training Loss.}
\textbf{Overall Training Loss.}
With Eq.~(\ref{eequa7}), Eq.~(\ref{eequa8}) and Eq.~(\ref{eequa9}), the final loss of our BP-triplet Net is given by
\setlength\abovedisplayskip{2pt}
\setlength\belowdisplayskip{2pt}
\begin{equation}
\begin{split}
%{\mathcal{L}}={{\mathcal{L}}_{adv}}+{{\mathcal{L}}_{{\rm{F-tri}}}}+{{\mathcal{L}}_{c}},
{\mathcal{L}}=\lambda_1 {{\mathcal{L}}_{adv}}+\lambda_2{{\mathcal{L}}_{{\rm{BP-tri}}}}+{{\mathcal{L}}_{c}},
\end{split}
\label{eequa10}
\end{equation}
where $\lambda_1$ and $\lambda_2$ are hyper-parameters.

The optimization problem is to find the parameters $\mathop {\theta _f }\limits^ \wedge,\mathop {\theta _y }\limits^ \wedge$ and $\mathop {\theta _d }\limits^ \wedge $. There is
\setlength\abovedisplayskip{2pt}
\setlength\belowdisplayskip{2pt}
\begin{equation}
\begin{array}{l}
 \mathop {(\theta _f }\limits^\wedge,\mathop {\theta_y }\limits^\wedge )= \arg\mathop {\min}\limits_{\theta_f,\theta_y }{\mathcal{L}}(\theta_f,\theta_y,\theta_d) \\
 \mathop {(\theta _d }\limits^\wedge)= \arg\mathop{\max }\limits_{\theta _d }{\mathcal{L}}(\theta_f,\theta _y,\theta_d). \\
 \end{array}
\label{eequa11}
\end{equation}

The optimization procedure is following the basic CNN protocol, since the gradients in Eq.~(\ref{eequa11}) are computable. In our model, no extra CNN variables are introduced. The CNN network parameters can be solved with standard mini-batch SGD. The pseudo-label $\hat{y}^t_i$ of $x^t_i$ based on maximum posterior probability using CNN softmax classifier is progressively updated during optimization. The class distribution discrepancy can be better interpreted with the increasing confidence of the pseudo target labels.

\subsection{Implementation Details}
%\textbf{Implementation Details.}
Negative transfer may happen when the corresponding modes of the distributions across domains are falsely aligned. To promote positive transfer and combat negative transfer, we should find a technology to reveal the multimode structures underlying distributions on which BP-triplet loss can be performed.
In order to ease the domain bias, we adopt a two-step strategy and the algorithm of our model is summarized in \textbf{Algorithm 1}.
\begin{table}%\small%\footnotesize
\begin{tabular}{l}
\toprule
\textbf{Algorithm 1}  The Proposed BP-Triplet Net\\
\midrule
\textbf{Input:} Labeled source data ~$\mathcal{D}_s=\{(x_i^s,y_i^s)\}_{i=1}^{n_{s}}$, \\
\hspace*{0.9cm} Unlabeled target data ~$\mathcal{D}_t=\{x_j^t\}_{j=1}^{n_{t}}$,\\
\hspace*{0.9cm} Threshold $T$, max number of steps ($S_0$), \\
\hspace*{0.9cm} The number of samples ($N_0$).\\
\textbf{Procedure:}\\
1. Pre-train a classifier:\\  %\textbf{Step1:}
\hspace*{0.9cm} Train classifier by Eq.~(\ref{eequa8}) and Eq.~(\ref{eequa9}). \\
2.  \textbf{for} $s=1$; $s \leq S_0$; $s ++$ \textbf{do}\\ %\textbf{Step2:}
\hspace*{0.9cm}  if~~ $s ~\% ~2000 ==0 $ \\
\hspace*{1.1cm} 1). Use the classifier to assign pseudo label  \\
\hspace*{1.6cm} $\hat{\bm{y}}^t_i$ for target sample $\bm{x}^t_i$. \\
\hspace*{1.1cm} 2). Select pseudo labeled samples with \\
\hspace*{1.6cm} predicted score above threshold $T$.\\
\hspace*{0.9cm}  if~~ The sample number of the $c^{th}$ class $\geq N_0$  \\
\hspace*{1.1cm} 1). The samples in class $c$ are used for pairing.\\
\hspace*{1.1cm} 2). Train model by Eq.~(\ref{eequa10}). \\
~~~~\textbf{end for}\\
\textbf{Output:} Predicted class of $\bm{x}_j^t$.\\
\bottomrule
\end{tabular}
\end{table}

\textbf{Step 1:} We progressively sample data from both source and target domains, and such learning can help our model capture more accurate structural information of target data. Because the target labels are unavailable, we naturally assign them the pseudo labels predicted by source classifier. However, the false pseudo labels may deteriorate domain adaptation, so we employ some mechanisms to ensure the training effectiveness. On one hand, we get the target labels after several iterations of training. This procedure improves the performance of source classifier on target data, so that more accurate pseudo labels can be obtained. On the other hand, the image with high confidence probability has higher reliability intuitively. Thus, we set a threshold $T$ and only when the predicted score of a sample is above the threshold $T$, the sample can be selected for constructing triplet pairs.
%In this paper, we empirically set the threshold $T$ as a constant $T=0.9$. Additionally, in order to implement the class alignment, we select the pseudo labeled samples for triplet pairing only if the sample number of the class is larger than a constant $N_0$. In this paper, $N_0=3$.

Noteworthily, in this paper, we set a self adjusted dynamical threshold $T$.  The information entropy can reflect the predicted uncertainty through these predicted probabilities of the sample. The entropy of the ground-truth class exploits mostly the information from the ground-truth class. However, the information from the incorrect classes has been largely ignored. We aim to utilize the $T$ to exploit the information from the predicted ground-truth class and other incorrect classes. For the predicted ground-truth classes, we employ the entropy to minimize the uncertainty, while for those incorrect classes, we want to neutralize their predicted probabilities to relieve the influence to the predicted ground-truth class. That is, we want to maximize the information entropy to neutralize those incorrect class probabilities.
In order to dynamically select the pseudo labeled triplet samples, on one hand, we add a loss to maximize the entropy from the incorrect classes and minimize the entropy from the predicted classes by the information entropy. On the other, with the increasing of iterations, the pseudo labels are more precise, the information entropy of the predicted ground-truth class is lower intuitively. We define the $T$ to measure the uncertainty by comparing the entropy from the predicted ground-truth class and other incorrect classes. Additionally, in order to reduce noise, we set a warm start $T$ which is larger than $0.9$.
The $T$ is defined as follows and it becomes larger with the iterations:
\setlength\abovedisplayskip{2pt}
\setlength\belowdisplayskip{2pt}
\begin{equation}%\small
\begin{array}{l}
T= max(0.9, 1-\frac{-p_i \log (p_i)}{-\sum\limits_{c = 1}^{C } p_c \log (p_c)}), \\
 \end{array}
\label{eequaT}
\end{equation}
where $C$ is the number of classes and $p_c$ is the probability of predicting a sample to class $c$. The $p_i$ is the predicted ground-truth class probability of the sample.
If the predicted score  $p_i\geq T$, the pseudo label $i$ can be regarded as the true label for triplet pairs.

\textbf{Step 2:} After selection, we leverage the labeled source samples and pseudo labeled target samples to achieve class-wise alignment by similarity learning.  We follow the random sampling strategy in~\cite{Deng2018Domain}. In our method, one half samples come from source domain and the other half from target domain, for triplet pairing, \textit{e.g.}, ($x_a,x_p,x_n$). Then, with the CNN framework, the model in Eq.~(\ref{eequa10}) is trained. We set the focusing parameter $\gamma=1$ and the parameter $\alpha=1$ in exponential distribution. These simple experimental setting can avoid over-adjusting parameters and highlight the effectiveness of the method itself.

\subsection{Theoretical Analysis}
%\textbf{Theoretical Analysis.}%,Ben2010A
In this section, we follow~\cite{Chen2019PFAN} to analyze our method and we make use of the theory in domain adaptation~\cite{Ben2006Analysis} to exhibit that our approach improves the boundary of the expected error on the target samples.
Formally, let $\mathcal{H}$ be the hypothesis class, $\mathcal{S}$ and $\mathcal{T}$ represent two domains. The probabilistic boundary of the error of hypothesis ${h}$ on the target domain is defined as,
\begin{equation}%\small
\begin{split}
\forall h \in \mathcal{H},R_\mathcal{T} (h) \le R_\mathcal{S} (h) + \frac{1}{2}d_{\mathcal{H}\Delta \mathcal{H}} (\mathcal{S},\mathcal{T}) + \lambda.
\end{split}
\label{eequa12}
\end{equation}

Obviously, the expected error on the target samples, $R_\mathcal{T}(h)$, is bounded by three terms: (1) the expected error on the source domain, $R_\mathcal{S}(h)$; (2) $d_{\mathcal{H}\Delta \mathcal{H}} (\mathcal{S},\mathcal{T})$ is the domain divergence measured by a discrepancy distance between two distributions $\mathcal{S}$ and $\mathcal{T}$ \textit{w.r.t.} a hypothesis set $\mathcal{H}$; (3) the shared error of the ideal joint hypothesis, $\lambda$.

In Inequality.~(\ref{eequa12}), the first item, $R_\mathcal{S} (h)$ is expected to be small and prone to be optimized by a deep network since we have source labels. The second item, $d_{\mathcal{H}\Delta \mathcal{H}} (\mathcal{S},\mathcal{T})$, in our method, is minimized in the adversarial DA efforts by the domain discriminator.

However, it is not enough that a small $R_\mathcal{S}(h)$ and a small $d_{\mathcal{H}\Delta \mathcal{H}} (\mathcal{S},\mathcal{T})$ do not guarantee small $R_\mathcal{T}(h)$. It is possible
that $\lambda$ tends to be large when the cross-domain category alignment is not be explicitly enforced. (\textit{i.e.}, the marginal distribution is well aligned, but the class conditional distribution is not guaranteed).
Therefore, $\lambda$ needs to be bounded as well. We cannot directly measure $\lambda$ due to the absence of target true labels. However, we can resort the pseudo-labels instead of giving the approximate evaluation and minimization in our method.

Benefitting from our pseudo-labeled strategy, we follow~\cite{Chen2019PFAN} to show the \textbf{Theorem 1}:
%to analyze our method. As is shown in \textit{Theorem 1}, we have

We define $G_\mathcal{S}$ and $G_\mathcal{T}$ as the labeling functions for the source and target domains,  $R_\mathcal{T'}(\cdot)$ is the expected risk on the selected pseudo-labeled target set $\hat{\mathcal{D}}_t$.

\textbf{Theorem 1}. \textit{Let $G_\mathcal{\hat{T}}$ be the pseudo-labeling function, $R_\mathcal{T'} (G_\mathcal{S}, G_\mathcal{\hat{T}})$ and $R_\mathcal{T'} (G_\mathcal{T}, G_\mathcal{\hat{T}})$ represent the minimum shared error and the degree to which the target samples are falsely labeled on $\hat{\mathcal{D}}_t$, respectively. Then}
\begin{equation}\small
\begin{split}
\lambda \le \mathop {\min } \limits_{h \in \mathcal{H}} R_\mathcal{S} (h, G_\mathcal{S}) + R_\mathcal{T'} (h, G_\mathcal{\hat{T}} )+ 2R_\mathcal{T'} (G_\mathcal{S}, G_\mathcal{\hat{T}} )+ R_\mathcal{T'} (G_\mathcal{T}, G_\mathcal{\hat{T}} ),
\end{split}
\label{eequa13}
\end{equation}

It is easy to find a suitable $h$ in $\mathcal{H}$ to approximate the $G_\mathcal{S}$ and $G_\mathcal{\hat{T}}$ since we have the source labels and target pseudo-labels to train the classifier. However, on the one hand, we assume that when the category alignment has not been achieved, there exists an optimality gap between $G_\mathcal{S}$ and $G_\mathcal{\hat{T}}$.
On the other, as the pseudo-labels are not always correct, the gap between $G_\mathcal{{T}}$ and $G_\mathcal{\hat{T}}$ cannot be eliminated. Therefore, it is not enough that only consider the expected risk of $R_\mathcal{S} (h, G_\mathcal{S})$, as it may lead to underfitting or overfitting for target samples.

Regarding our method, in the following we will remark that our approach improves the boundary of the shared error of the joint hypothesis $\lambda$, such that the boundary of the expected error of target samples $R_\mathcal{T}(h)$ is improved.
\begin{itemize}
\item Minimizing $R_\mathcal{S} (h, G_\mathcal{S}) + R_\mathcal{T'} (h, G_\mathcal{\hat{T}})$.

The cross-entropy loss in source domain is utilized to train the classifier. To alleviate the overfitting of source domain~(\textit{i.e.}, a non-saturated source classifier), target samples should be joint to train a generalized model. In our method, pseudo-labeled target samples are additionally leveraged to train the domain-invariant features. Additionally, leveraging the BP-triplet loss, cross-domain category distributions are well aligned, and the aforementioned optimality gap is removed and the generalized classifier can achieve a better performance, \textit{i.e.}, a smaller
$R_\mathcal{S} (h, G_\mathcal{S}) + R_\mathcal{T'} (h, G_\mathcal{\hat{T}})$.

%joint marginal and conditional distribution is leveraged by the Eq.~(\ref{eequa3}) and Eq.~(\ref{eequa5}). Considering that $f_\mathcal{S}$ is adopted to replace $f_\mathcal{\hat{T}}$ in our approach,
%the joint distribution strategy guides the generalized classifier to a better performance, \textit{i.e.}, a smaller Eq.~(\ref{eequa6})
%$R_\mathcal{S} (h, f_\mathcal{S}) + R_\mathcal{T'} (h, f_\mathcal{\hat{T}})$.

%The proposed method considers joint marginal and conditional distribution, and the features in both domain level and class level are aligned. Thus, the cross-domain category distributions can be well aligned. The optimality gap between $f_\mathcal{S}$ and $f_\mathcal{\hat{T}}$ is decreased and this guides the adaptation model to a better target performance, \textit{i.e.}, a smaller Eq.~(\ref{eequa6})
%$R_\mathcal{S} (h, f_\mathcal{S}) + R_\mathcal{T'} (h, f_\mathcal{\hat{T}})$.

\item Minimizing the shared error $R_\mathcal{T'} (G_\mathcal{S}, G_\mathcal{\hat{T}} )$.

\textit{By considering the 0-1 loss function $\sigma$ for $R_\mathcal{T'}$,we have}
\begin{equation}%\small
\begin{split}
&R_{T'} (G_S ,G_\mathcal{\hat{T}})=E_{x\sim T'}[\sigma (G_S (f(x)),G_\mathcal{\hat{T}} (f(x)))] \\
&= E_{x\sim T'} [|\sigma (G_S (f(x)),y_1 ) - \sigma (G_\mathcal{\hat{T}} (f(x)),y_2 )|],\\
&where\\
&|\sigma (G_S (f(x)),y_1 ) - \sigma (G_\mathcal{\hat{T}} (f(x)),y_2 )| = \left\{ {\begin{array}{*{3}c}
   1 & {if~ y_1  \ne y_2 }  \\
   0 & {if~ y_1  = y_2 }.  \\
\end{array}} \right.
\end{split}
\label{eequa14}
\end{equation}

The Eq.~(\ref{eequa14}) term aims to achieve class-level alignment cross domains. Considering the relation with our method, on one hand, by Eq.~(\ref{eequa7}) of triplet loss part, the class-level aligned features are trained. On the other, by Eq.~(\ref{eequa7}) of weighted part and Eq.~(\ref{eequa9}), cross-domain pairs are considered in our method, and better domain-level alignment can be achieved. In this way, the $k_{th}$ class in source domain $\hat{\mathcal{D}}_s^k$ and the same pseudo-labeled target class $\hat{\mathcal{D}}_t^k$ are aligned. When the categories have been better clustered, it is safe to assume that $y_1  = y_2$. Thus, $R_\mathcal{T'} (G_\mathcal{S}, G_\mathcal{\hat{T}} )$ is expected to be minimized.

%\textit{Our approach aims to collaboratively align global-level and category-level features, \textit{i.e.}, it aligns the $k_{th}$ class in source domain $\hat{\mathcal{D}}_s^k$ with the same pseudo-labeled target class $\hat{\mathcal{D}}_t^k$.}

\item Minimizing the degree to which the target samples are falsely labeled on $\hat{\mathcal{D}}_t$: $R_\mathcal{T'} (G_\mathcal{T}, G_\mathcal{\hat{T}} )$.

In order to reduce the gap between $G_\mathcal{T}$ and $G_\mathcal{\hat{T}}$, our method adopts the co-training strategy. The reliable pseudo-labeled target samples are progressively selected during the optimization. Then $R_\mathcal{T'} (G_\mathcal{T}, G_\mathcal{\hat{T}} )$ is minimized in the procedure.

\end{itemize}

\begin{figure}[t]
\begin{center}
  \includegraphics[width=1.0\linewidth]{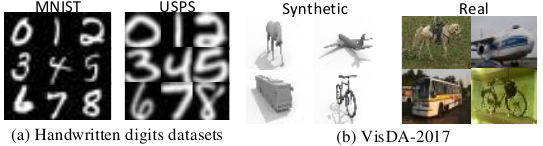}
\end{center}
  \setlength{\abovecaptionskip}{0pt}
%\captionsetup{justification=centering}
   \caption{Some examples from handwritten digits datasets.}
   \label{mnist}
\end{figure}

\begin{table}[t]\normalsize
\begin{center}
\setlength{\tabcolsep}{1.5mm}{
\begin{tabular}{  l | c c   |c}
\toprule
Handwritten&M $\to$ U&U $\to$ M&Avg. \\
\hline
\textit{ADDA}~\cite{Tzeng2017Adversarial}&$89.4$&$90.1$&$89.8$\\
\hline
\textit{CoGAN}~\cite{liu2016coupled}&$95.6$ &$93.1$ &$94.3$\\
\hline
\textit{UNIT}~\cite{Liu2017UNIT}&$\bf{96.0}$ &$93.6$ &$94.8$\\
\hline
\textit{CDAN}~\cite{Long2017Conditional}&$93.9$ &$96.9$ &$95.4$ \\
\hline
\textit{CYCADA}~\cite{Hoffman2017CyCADA}&$95.6$ &$96.5$ &$\bf{96.1}$\\
\hline
\textit{Ours}&$94.1$ &$\bf{98.0}$ &$\bf{96.1}$\\
\bottomrule
\end{tabular}}
%\end{tabularx}
\end{center}
\setlength{\abovecaptionskip}{0pt}
%\captionsetup{justification=centering}
\caption{Recognition accuracies ($\%$)  on Office-Home dataset. All models utilize LeNet as base architecture.}
\label{tabmnist}
\end{table}

\section{Experiment}
\begin{figure}[t]
\begin{center}
  \includegraphics[width=1\linewidth]{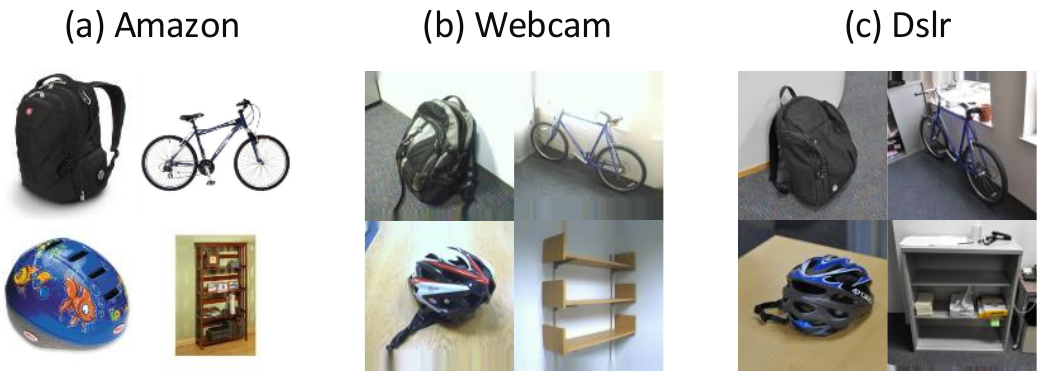}
\end{center}
  \setlength{\abovecaptionskip}{0pt}
%\captionsetup{justification=centering}
   \caption{Some examples from Office-31 object dataset.}
   \label{office31}
\end{figure}

\begin{figure}[t]
\begin{center}
  \includegraphics[width=1\linewidth]{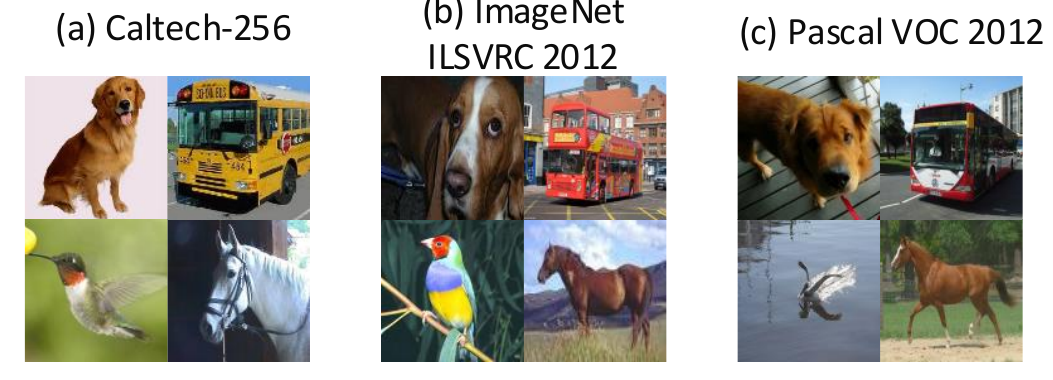}
\end{center}
  \setlength{\abovecaptionskip}{0pt}
%\captionsetup{justification=centering}
   \caption{Some examples from ImageCLEF-DA image dataset.}
   \label{imageclef}
\end{figure}

\begin{figure}[t]
\begin{center}
  \includegraphics[width=1\linewidth]{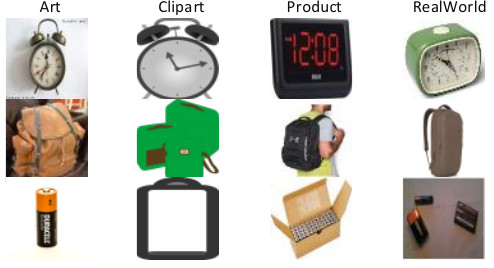}
\end{center}
  \setlength{\abovecaptionskip}{0pt}
%\captionsetup{justification=centering}
   \caption{Some examples from Office-Home dataset.}
   \label{officehome}
\end{figure}

\begin{table*}[t] \small   %\footnotesize
\begin{center}
\setlength{\tabcolsep}{3.5mm}
{
%\newcolumntype{Y}{>{\centering\arraybackslash}X}
%\begin{tabularx}{\textwidth}{|c|*{8}{Y|}}\hline
\begin{tabular}{  l | c c  c  c  c c  |c}
%\toprule
\hline
Office-31&A$\to$W&D$\to$W&W$\to$D&A$\to$D&D$\to$A&W$\to$A&Avg. \\
\hline
\textit{Source Only}&$68.4$&$96.7$&$99.3$&$68.9$&$62.5$&$60.7$&$76.1$\\
\hline
\textit{TCA}~\cite{pan2011domain}&$72.7$ &$96.7$ &$99.6$ &$74.1$ &$61.7$ &$60.9$ &$77.6$\\
\textit{GFK}~\cite{Gong2012Geodesic}&$72.8$ &$95.0$ &$98.2$ &$74.5$ &$63.4$ &$61.0$ &$77.5$\\
\textit{DDC}~\cite{tzeng2014deep}&$75.6$ &$96.0$ &$98.2$ &$76.5$ &$62.2$ &$61.5$ &$78.3$\\
\textit{DAN}~\cite{long2015learning}&$80.5$ &$97.1$ &$99.6$ &$78.6$ &$63.6$ &$62.8$ &$80.4$\\
\textit{RTN}~\cite{long2016unsupervised}&$84.5$ &$96.8$ &$99.4$ &$77.5$ &$66.2$ &$64.8$ &$81.6$\\
\textit{DANN}~\cite{ganin2017domain}&$82.0$ &$96.9$ &$99.1$ &$79.7$ &$68.2$ &$67.4$ &$82.2$\\
\textit{ADDA}~\cite{Tzeng2017Adversarial}&$86.2$ &$96.2$ &$98.4$ &$77.8$ &$69.5$ &$68.9$ &$82.9$\\
\textit{JAN}~\cite{long2017deep}&$85.4$ &$97.4$ &$99.8$ &$84.7$ &$68.6$ &$70.0$ &$84.3$\\
\textit{MADA}~\cite{Cao2018Partial}&$90.0$ &$97.4$ &$99.6$ &$87.8$ &$70.3$ &$66.4$ &$85.2$\\
%\textit{CAN}&$81.5$ &$98.2$ &$99.7$ &$85.5$ &$65.9$ &$63.4$ &$82.4$\\
%\textit{EM}&$86.8$ &$\bf{99.3}$ &$\bf{100.0}$&$87.2$ &$71.2$ &$71.8$ &$86.1$\\
%\textit{SimNet}&$88.6$ &$98.2$ &$99.7$ &$85.3$ &$73.4$ &$71.8$ &$86.2$\\
\textit{GTA}~\cite{Sankaranarayanan2017Generate}&$89.5$ &$97.9$ &$99.8$ &$87.7$ &$72.8$ &$71.4$ &$86.5$\\
\textit{MCD}~\cite{saito2017maximum}&$88.6$ &$98.5$ &$\bf{100.0}$ &$92.2$ &$69.5$ &$69.7$ &$86.5$\\
\textit{SAFN+ENT}~\cite{xu2019larger}&$90.1$ &$98.6$ &${99.8}$ &$90.7$ &$73.0$ &$70.1$ &$87.1$\\
\textit{iCAN}~\cite{zhang2018collaborative}&$92.5$ &${98.8}$ &$\bf{100.0}$ &$90.1$ &$72.1$ &$69.9$ &$87.2$\\
\textit{CDAN}~\cite{Long2017Conditional}&$94.1$ &$98.6$ &$\bf{100.0}$ &$92.9$ &$71.0$ &$69.3$ &$87.7$\\
\textit{TADA}~\cite{TADA2019}&$94.3$ &$98.7$ &$99.8$ &$91.6$ &$72.9$ &$73.0$ &$88.4$\\
\textit{SymNet}~\cite{Zhang2019Domain}&$90.8$ &${98.8}$ &$\bf{100.0}$ &${93.9}$ &${74.6}$ &$72.5$ &$88.4$\\
\textit{ALDA}~\cite{2020Adversarial}&$95.6$ &${97.7}$ &$\bf{100.0}$ &$\bf{94.0}$ &${72.2}$ &$72.5$ &$88.7$\\
\textit{MDD+IA}~\cite{jiang2020implicit}&$90.3$ &${98.7}$ &${99.8}$ &${92.1}$ &$\bf{75.3}$ &$74.9$ &$88.8$\\
\textit{DADA}~\cite{tang2020discriminative}&$92.3$ &$\bf{99.2}$ &$\bf{100.0}$ &${93.9}$ &${74.4}$ &$74.2$ &$89.0$\\
\textit{BCDM}~\cite{li2020bi}&$95.4$ &${98.6}$ &$\bf{100.0}$ &${93.8}$ &${73.1}$ &$73.0$ &$89.0$\\
\hline
\textit{Ours}&$\bf{96.7}$ &$98.4$ &$\bf{100.0}$ &$91.7$ &$73.1$ &$\bf{75.0}$ &$\bf{89.2}$\\
\hline
\end{tabular}
%\end{tabularx}
}
%\bf{
\end{center}
\setlength{\abovecaptionskip}{0pt}
%\captionsetup{justification=centering}
\caption{ Recognition accuracies ($\%$) on the Office31 dataset.
All models utilize ResNet-50 as base architecture.
}
\label{tab1}
\end{table*}

\begin{table*}[t]\small
\begin{center}
\setlength{\tabcolsep}{3.5mm}
{
\begin{tabular}{  l | c c  c  c  c c  |c}
\toprule
ImageCLEF-DA&I$\to$P&P$\to$I&I$\to$C&C$\to$I&C$\to$P&P$\to$C&Avg. \\
\hline
\textit{Source Only}&$74.8$&$83.9$&$91.5$&$78.0$&$65.5$&$91.2$&$80.7$\\
\hline
\textit{DDC}~\cite{tzeng2014deep}&$74.6$ &$85.7$ &$91.1$ &$82.3$ &$68.3$ &$88.8$ &$81.8$\\
\textit{DAN}~\cite{long2015learning} &$74.5$ &$82.2$ &$92.8$ &$86.3$ &$69.2$ &$89.8$ &$82.5$\\
\textit{RTN}~\cite{long2016unsupervised}&$75.6$ &$86.8$ &$95.3$ &$86.9$ &$72.7$ &$92.2$ &$84.9$\\
\textit{DANN}~\cite{ganin2017domain}&$75.0$ &$86.0$ &$96.2$ &$87.0$ &$74.3$ &$91.5$ &$85.0$\\
\textit{JAN}~\cite{long2017deep}&$76.8$ &$88.0$ &$94.7$ &$89.5$ &$74.2$ &$91.7$ &$85.8$\\
%\textit{CAN} &$78.2$ &$87.5$ &$94.2$ &$89.5$ &$75.8$ &$89.2$ &$85.7$\\
%\textit{CDAN}&$76.7$ &$90.6$ &$97.0$ &$90.5$ &$74.5$ &$93.5$ &$87.1$\\
\textit{MADA}~\cite{Cao2018Partial}&$75.0$ &$87.9$ &$96.0$ &$88.8$ &$75.2$ &$92.2$ &$85.8$\\
\textit{iCAN}~\cite{zhang2018collaborative} &$\bf{79.5}$ &$89.7$ &$94.7$ &$89.9$ &$\bf{78.5}$ &$92.0$ &$87.4$\\
\textit{CDAN}~\cite{Long2017Conditional}&$77.7$ &$90.7$ &$\bf{97.7}$ &$\bf{91.3}$ &$74.2$ &${94.3}$ &$87.7$\\
\hline
\textit{Ours}&$78.6$&$\bf{92.1}$&$96.6$&$\bf{91.3}$&$77.8$&$\bf{96.1}$&$\bf{88.8}$\\
\bottomrule
\end{tabular}}
%\end{tabularx}
\end{center}
\setlength{\abovecaptionskip}{0pt}
%\captionsetup{justification=centering}
\caption{Recognition accuracies ($\%$) on ImageCLEF-DA.
All models utilize ResNet-50 as base architecture.
}
\label{tab2}
\end{table*}

\subsection{Datasets and Baselines}
In this section, several benchmark datasets, \textit{i.e.}, not only the toy datasets as USPS+MNIST datasets, but also Office-31 dataset~\cite{saenko2010adapting}, ImageCLEF-DA~\cite{long2017deep} dataset\footnote{\url{http://imageclef.org/2014/adaptation}}, Office-Home~\cite{venkateswara2017deep} dataset\footnote{\url{http://hemanthdv.org/OfficeHome-Dataset}} and VisDA-2017~\cite{2020Adversarial}  dataset\footnote{\url{1https://github.com/VisionLearningGroup/taskcv-2017-public/}} are adopted for evaluation.

In our experiment, the target labels are unseen by following the standard evaluation protocol of UDA~\cite{long2017deep}. Our implementation is based on the PyTorch framework. For the toy datasets of handwritten digit datasets, we utilize the LeNet as the backbone network, while for other datasets except VisDA-2017, we use the pre-trained ResNet-50 for fairness. In VisDA-2017 dataset, following other compared methods, we adopted the pre-trained ResNet-101 as the backbone network.
We fine-tune convolutional layers, and apply back-propagation to train the classifier layer and domain discriminator. Whatever module trained from scratch, its learning rate was set to be 10 times that of the lower layers. We adopt mini-batch stochastic gradient descent (SGD)
with momentum of 0.9 using the learning rate and progressive training strategies as in CDAN~\cite{Long2017Conditional}. In the process of selecting pseudo-labeled samples, the threshold $T$ is dynamically adjusted.
%as backbone network and re-train the parameters of high-level layers
In our paper, we fix the model hyper-parameters $\lambda_1=1$ and $\lambda_2=1$ throughout all experiments.
For the BP-triplet loss, we set the margin $m$ and $N_0$ in triplet loss as 0.3 and 3 following the setting as usual, respectively.
%For the $K-$value in the classification loss of Eq.~(\ref{eequa14}), we choose $K=3$ for Office-31 and ImageCLEF-DA datasets, and $K=C/2$ in Office-Home dataset as this dataset has a variety of categories and large domain gap.
%$C$ is the number of categories.

Rank-1 classification accuracy is adopted  for comparison. For toy datasets of handwritten digit, as there are plenty of different configutations, we only show some the recent results with the same backbone and training/test split for fair comparison. We compared with  ADDA~\cite{Tzeng2017Adversarial}, CoGAN~\cite{liu2016coupled}, UNIT~\cite{Liu2017UNIT},  CYCADA~\cite{Hoffman2017CyCADA} and CDAN~\cite{Long2017Conditional}.
For other datasets, our compared baseline methods include DAN~\cite{long2015learning}, DANN~\cite{ganin2017domain}, JAN~\cite{long2017deep}, CDAN~\cite{Long2017Conditional}. Besides, on Office-31 dataset, we compare with TCA~\cite{pan2011domain}, GFK~\cite{Gong2012Geodesic}, DDC~\cite{tzeng2014deep}, RTN~\cite{long2016unsupervised}, ADDA~\cite{Tzeng2017Adversarial}, MADA~\cite{Cao2018Partial}, GTA~\cite{Sankaranarayanan2017Generate}, MCD~\cite{saito2017maximum}, iCAN~\cite{zhang2018collaborative}, TADA~\cite{TADA2019}, SymNet~\cite{Zhang2019Domain} and so on. On ImageCLEF-DA dataset, RTN~\cite{long2016unsupervised}, MADA~\cite{Cao2018Partial} and iCAN~\cite{zhang2018collaborative} are compared.
On OfficeHome dataset, TADA~\cite{TADA2019} and SymNet~\cite{Zhang2019Domain}, ALDA~\cite{2020Adversarial}, SAFN~\cite{xu2019larger} and ATM~\cite{li2020maximum} are compared.
On VisDA-2017 dataset,  ADR~\cite{2017Adversarial}, ALDA~\cite{2020Adversarial}, SAFN~\cite{xu2019larger} and SWD~\cite{2019Sliced} and are compared.
The results of the compared methods were from the original papers.
Noteworthily, the results of the compared methods were from the original papers. The backbone network is the same for fairness.

\begin{table*}[t]\small    %\footnotesize
\begin{center}
%\newcolumntype{Y}{>{\centering\arraybackslash}X}
\setlength{\tabcolsep}{1.0mm}{
%\resizebox{\textwidth}{15mm}{
\begin{tabular}{  l | c c  c  c  c c c c  c  c  c c  |c}
\toprule
OfficeHome&Ar$\to$Cl&Ar$\to$Pr&Ar$\to$Rw&Cl$\to$Ar&Cl$\to$Pr&Cl$\to$Rw&Pr$\to$Ar&Pr$\to$Cl&Pr$\to$Rw&Rw$\to$Ar&Rw$\to$Cl&Rw $\to$Pr&Avg. \\
\hline
\textit{Source Only}&$34.9$&$50.0$&$58.0$&$37.4$&$41.9$&$46.2$&$38.5$&$31.2$&$60.4$&$53.9$&$41.2$&$59.9$&$46.1$\\
\hline
\textit{DAN}~\cite{long2015learning}&$43.6$ &$57.0$ &$67.9$ &$45.8$ &$56.5$ &$60.4$ &$44.0$&$43.6$ &$67.7$ &$63.1$ &$51.5$ &$74.3$ &$56.3$\\
\textit{DANN}~\cite{ganin2017domain}&$45.6$ &$59.3$ &$70.1$ &$47.0$ &$58.5$ &$60.9$ &$46.1$&$43.7$ &$68.5$ &$63.2$ &$51.8$ &$76.8$ &$57.6$\\
\textit{JAN}~\cite{long2017deep}&$45.9$ &$61.2$ &$68.9$ &$50.4$ &$59.7$ &$61.0$ &$45.8$&$43.4$ &$70.3$ &$63.9$ &$52.4$ &$76.8$ &$58.3$\\
\textit{CDAN}~\cite{Long2017Conditional}&$50.7$ &$70.6$ &$76.0$ &$57.6$ &$70.0$ &$70.0$ &$57.4$&$50.9$ &$77.3$ &$70.9$ &$56.7$ &$81.6$ &$65.8$\\
\textit{ALDA}~\cite{2020Adversarial}&$53.7$ &$70.1$ &$76.4$ &$60.2$ &$72.6$ &$71.5$ &$56.8$&$51.9$ &$77.1$ &$70.2$ &$56.3$ &$82.1$ &$66.6$\\
\textit{SAFN}~\cite{xu2019larger}&$52.0$ &$71.7$ &$76.3$ &$64.2$ &$69.9$ &$71.9$ &$63.7$&$51.4$ &$77.1$ &$70.9$ &$57.1$ &$81.5$ &$67.3$\\
\textit{TADA}~\cite{TADA2019}&$53.1$ &$72.3$ &$77.2$ &$59.1$ &$71.2$ &$72.1$ &$59.7$&$53.1$ &$78.4$ &$72.4$ &$60.0$ &$82.9$ &$67.6$\\
\textit{SymNet}~\cite{Zhang2019Domain}&$47.7$ &$72.9$ &$\bf{78.5}$ &$\bf{64.2}$ &$71.3$ &$\bf{74.2}$ &$\bf{64.2}$&$48.8$ &$\bf{79.5}$ &$\bf{74.5}$ &$52.6$ &$82.7$ &$67.6$\\
\textit{ATM}~\cite{li2020maximum}&$52.4$ &$72.6$ &$78.0$ &$61.1$ &$\bf{72.0}$ &$72.6$ &$59.5$&$52.0$ &$79.1$ &$73.3$ &$58.9$ &$83.4$ &$67.9$\\
\hline
\textit{Ours}&$\bf{55.1}$&$\bf{74.1}$&$77.5$&$60.3$&$\bf{72.0}$&$71.4$&
$58.8$&$\bf{52.5}$&$79.0$&$70.6$&$\bf{62.0}$&$\bf{83.6}$&$\bf{68.1}$\\
\bottomrule
\end{tabular}}
%\end{tabularx}
\end{center}
\setlength{\abovecaptionskip}{0pt}
\captionsetup{justification=centering}
\caption{Recognition accuracies ($\%$)  on Office-Home dataset. All models utilize ResNet-50 as base architecture.}
\label{tab3}
\end{table*}

\begin{table*}[t]\small    %\footnotesize
\begin{center}
%\newcolumntype{Y}{>{\centering\arraybackslash}X}
\setlength{\tabcolsep}{1.3mm}{
%\resizebox{\textwidth}{15mm}{
\begin{tabular}{  l | c c  c  c  c c c c  c  c  c c  |c}
\toprule
VisDA&airplane&bicycle&bus&car&horse&knife&motorcycle&person&plant&skateboard&train&truck &Avg. \\
\hline
\textit{Source Only}&$55.1$&$53.3$&$61.9$&$59.1$&$80.6$&$17.9$&$79.7$&$31.2$&$81.0$&$26.5$&$73.5$&$8.5$&$52.4$\\

\textit{DANN}~\cite{ganin2017domain}&$81.9$ &$77.7$ &$82.8$ &$44.3$ &$81.2$ &$29.5$ &$65.1$&$28.6$ &$51.9$ &$54.6$ &$82.8$ &$7.8$ &$57.4$\\
\textit{DAN}~\cite{long2015learning}&$68.1$ &$15.4$ &$76.5$ &$87.0$ &$71.1$ &$48.9$ &$82.3$&$51.5$ &$88.7$ &$33.2$ &$88.9$ &$42.2$ &$62.8$\\
\textit{JAN}~\cite{long2017deep}&$75.7$ &$18.7$ &$82.3$ &$86.3$ &$70.2$ &$56.9$ &$80.5$&$53.8$ &$92.5$ &$32.2$ &$84.5$ &$54.5$ &$65.7$\\
\textit{MCD}~\cite{saito2017maximum}&$87.0$ &$60.9$ &$83.7$ &$64.0$ &$88.9$ &$79.6$ &$84.7$&$76.9$ &$88.6$ &$40.3$ &$83.0$ &$25.8$ &$71.9$\\
\textit{CDAN}~\cite{Long2017Conditional}&$85.2$ &$66.9$ &$83.0$ &$50.8$ &$84.2$ &$74.9$ &$88.1$&$74.5$ &$83.4$ &$76.0$ &$81.9$ &$38.0$ &$73.9$\\
\textit{ADR}~\cite{2017Adversarial}&$87.8$ &$79.5$ &$83.7$ &$65.3$ &$92.3$ &$61.8$ &$88.9$&$73.2$ &$87.8$ &$60.0$ &$85.5$ &$32.3$ &$74.8$\\
\textit{SAFN}~\cite{xu2019larger}&$93.6$ &$61.3$ &$\bf{84.1}$ &$\bf{70.6}$ &$\bf{94.1}$ &$79.0$ &$\bf{91.8}$&$79.6$ &$89.9$ &$55.6$ &$\bf{89.0}$ &$24.4$ &$76.1$\\
\textit{SWD}~\cite{2019Sliced}&$90.8$ &$\bf{82.5}$ &${81.7}$ &${70.5}$ &$91.7$ &${69.5}$ &${86.3}$&$77.5$ &${87.4}$ &${63.6}$ &$85.6$ &$29.2$ &$76.4$\\
\textit{ALDA}~\cite{2020Adversarial}&$93.8$ &$74.1$ &${82.4}$ &${69.4}$ &$90.6$ &${87.2}$ &${89.0}$&$67.6$ &$\bf{93.4}$ &${76.1}$ &${87.7}$ &$22.2$ &$77.8$\\
\hline
\textit{Ours}&$\bf{95.2}$&${75.5}$&$81.1$&$47.0$&${91.8}$&$\bf{96.6}$&
$89.0$&$\bf{83.1}$&$86.2$&$\bf{84.0}$&${87.3}$&$\bf{37.3}$&$\bf{79.5}$\\
\bottomrule
\end{tabular}}
%\end{tabularx}
\end{center}
\setlength{\abovecaptionskip}{0pt}
\captionsetup{justification=centering}
\caption{Recognition accuracies ($\%$)  on VisDA-2017 dataset. All models utilize ResNet-101 as base architecture.}
\label{tab4}
\end{table*}

\subsection{Comparisons with State-of-the-Arts}
\textbf{Results on Handwritten Digits Datasets}.
 MNIST~(\textbf{M}) and USPS~(\textbf{U}) datasets are toy handwritten digits datasets in domain adaptation. Some samples are shown in Figure~\ref{mnist}~(a). They are standard digit recognition datasets containing handwriten digits from $0-9$. Since the same digits across two datasets belong to different distributions, it is necessary to perform domain adaptation. MNIST consists of 60,000 training images and 10,000 test images of size $28 \times 28$.  USPS consists of 7,291 training images and 2,007 test images of size $16 \times 16$. We follow the experimental settings of~\cite{Hoffman2017CyCADA} to construct two tasks: $U \to M$ and $M \to U$. The experimental results are shown in Table~\ref{tabmnist}, from which we verify that our method achieves competitive performance in toy datasets.

\textbf{Results on Office-31 Dataset~\cite{saenko2010adapting}}.
This dataset is a challenging benchmark dataset for cross-domain object recognition. This dataset includes three domains such as Amazon (\textbf{A}), Webcam (\textbf{W}) and Dslr (\textbf{D}) as shown in Figure~\ref{office31}. It contains 4,652 images from 31 object classes. With each domain worked as source and target alternatively, 6 cross-domain tasks are formed, \textit{e.g.}, \textbf{A} $\to$ \textbf{D} ,\textbf{W} $\to$ \textbf{D}, \textit{etc}. In experiment, we follow the same experimental protocol as~\cite{long2015learning}. The recognition accuracy is reported in Table~\ref{tab1}.
For fair comparison, all models use the same ResNet-50 architecture.
From the results, we observe that our method ($89.2\%$ in average) outperforms state of the arts.
\textit{e.g.}, our accuracy exceeds the TADA~\cite{TADA2019} and SymNet~\cite{Zhang2019Domain} by 0.8\%.
%Noteworthily, it is very hard to have a promotion in this representative benchmark dataset.
This demonstrates that our model can effectively alleviate the model bias problem.

\textbf{Results on ImageCLEF-DA Dataset~\cite{long2017deep}}. This dataset is a benchmark for ImageCLEF 2014 domain adaptation challenge and some example images are displayed in Figure~\ref{imageclef}. It contains 12 common categories shared by three public datasets: Caltech-256 (\textbf{C}), ImageNet ILSVRC 2012 (\textbf{I}) and Pascal VOC 2012 (\textbf{P}). In each domain, there are 50 images per class and totally 600 images are constructed. We evaluate all methods across three transfer domains and build 6 cross-domain tasks: \textit{e.g.}, \textbf{I} $\to$ \textbf{P}, \textbf{P} $\to$ \textbf{I}, \textit{etc}. We compare our method with the baseline model (ResNet-50) and the existing deep domain adaptation methods.
The experimental results are shown in Table~\ref{tab2}, from which we observe that our method ($88.8\%$ in average) still outperforms other state-of-the-art methods.\textit{e.g.}, CDAN\cite{Long2017Conditional} method achieves accuracy of $87.7\%$, and we are able to outperform it by $1.1\%$.

\textbf{Results on Office-Home Dataset~\cite{venkateswara2017deep}}.
This is a challenging dataset  which consists of 15,500 images. There are 65 categories coming from four significantly different domains: Artistic images~(\textbf{Ar}), Clip Art~(\textbf{Cl}), Product images~(\textbf{Pr}) and Real-World images~(\textbf{Rw}). Each domain worked as source and target alternatively, there are 12 DA tasks constructed on this dataset.
Several example images are shown in Figure~\ref{officehome}.
We follow the same experimental protocol as~\cite{Long2017Conditional}, and compare against several recently reported results of well-known deep domain adaptation methods on the Office-Home dataset.
The results are shown in Table~\ref{tab3}, from which we observe that our method achieves the best performance compared with most state of the arts in average classification accuracy. Since domain alignment is category agnostic in previous work, it is possible that the aligned domains are not classification friendly in the presence of a number of categories. However, our similarity metric learning based class alignment model BP-triplet Net yields significant improvement on these more difficult DA tasks.

\textbf{Results on VisDA-2017 Dataset~\cite{peng2017visda}}.
This is a very large and challenging dataset for domain adaptation, which consists of more than 280K images from 12 categories. It comes from three different domains: a training domain (\textbf{Synthetic}), a validation
domain (\textbf{Real}) and a testing domain. With a training domain (\textbf{Synthetic}) and a validation
domain (\textbf{Real}) worked as source and target domain respectively, there is a Synthetic$\to$Real DA task constructed on this dataset.
Several example images are shown in Figure~\ref{mnist}~(b).

We follow the same experimental protocol as~\cite{Long2017Conditional}, and compare against several recently reported results of deep methods on the VisDA-2017 dataset.
The results are shown in Table~\ref{tab4}, from which we observe that our method achieves the competitive performance compared with most state of the arts in average classification accuracy.

\section{Discussion}

\subsection{Ablation Study}
%\textbf{Ablation Study.}
We propose to utilize metric learning to align class relations for alleviating the class bias problem. Table~\ref{tab4} presented the ablation analysis results under different model variants with some loss removed. The baseline of \textit{Source Only} represents that only the cross-entropy loss of source classifier is trained. \textit{Ours}~($w/o~\mathcal{L}_{adv}$) means that we use our method to train the network without adversarial loss $\mathcal{L}_{adv}$, and the performance is increased to 83.8\%. \textit{DANN} is another important baseline, in which the source classifier and domain alignment both are taken into account, and the performance is increased to 82.1\%.
Besides, the entropy minimization loss of target samples is optimized additionally in our method, in order to verify the proposed BP-triplet loss, we denote it as \textit{DANN~(Em)} and it also can be regarded as a baseline. From Table~\ref{tab4}, we can observe that the performance of our method is increased from 87.2\% to 90.9\% by jointly aligning domain (\textit{i.e.}, $\mathcal{L}_{adv}$) and class distribution (\textit{i.e.}, $\mathcal{L}_{BP-tri}$).
Our collaborative method combining class alignment and domain alignment achieves the best performance.

Specifically, our BP-triplet loss contributes two aspects: 1) triplet loss on sample pairs encourages the distance between positive pairs closer than the negative pairs by a margin; 2) the weight aims to down-weight the easy sample pairs and up-weight the hard pairs to give a strong enforcement for characterizing the pair-wise importance. Simultaneously, it enhances the domain confusion during the class-level alignment. For better insight into the importance of the proposed BP-triplet loss, we present Table~\ref{tab6} for the model ablation analysis.
Our BP-triplet loss is the variant of the standard triplet loss, \textit{i.e.}, \textit{DANN~(Em}$+{\mathcal{L}_{tri}}$). We give the comparison between the two losses to verify the effectiveness of the weight part. From the results, the performance is decreased from 90.9\% to 89.5\%  without weight part. The proposed BP-triplet loss is experimentally verified.
Additionally, the pseudo target labels strategy is also taken into consideration, for validating the triplet loss with target samples, we have demonstrated the effectiveness of pseudo labels in experiment. The performance is decreased from 89.5\% to 88.6\% after removing the target pseudo labels, \textit{i.e.}, \textit{DANN~(Em}$+\mathcal{L}_{tri-s}$).
By comparing the performance of different model variants, we find that the proposed model with BP-triplet loss and pseudo target labels can greatly boost the performance of unsupervised domain adaptation.

\subsection{Quantitative Distribution Discrepancy}
%\textbf{Quantitative Distribution Discrepancy}.
$\mathcal{A}$-distance \cite{ben2010theory} jointly formulates source and target risk, and we use $\mathcal{A}$-distance to measure the distribution discrepancy after domain adaptation.
It is defined as $d_\mathcal{A}=2(1-2\epsilon)$, where $\epsilon$ means the classification error of a binary domain classifier. From the expression, it is can be seen that with the increasing discrepancy between two domains, the error $\epsilon$ becomes smaller. Obviously, a large $\mathcal{A}$-distance denotes a large domain discrepancy. In our method, the distribution discrepancy analysis based on $\mathcal{A}$-distance in Office-31 dataset on tasks \textbf{A} $\to$ \textbf{W} and \textbf{W} $\to$ \textbf{D} is conducted by using \textit{ResNet}, \textit{DANN}, \textit{BP-Triplet Net} without adversarial learning~(\textit{Ours}~($w/o~\mathcal{L}_{adv}$)), and our \textit{BP-Triplet Net}, respectively.
Figure \ref{fig5}~(a) shows $\mathcal{A}$-distance on different tasks by using different models.
It is observed that $\mathcal{A}$-distance between domains after using our model is much smaller than that of other three methods, which suggests that our method is effective in reducing the domain discrepancy gap. Additionally, \textbf{W} $\to$ \textbf{D} has a much smaller $\mathcal{A}$-distance than \textbf{A} $\to$ \textbf{W} obviously.
From the classification accuracy in Table \ref{tab1}, the recognition rate of \textbf{W} $\to$ \textbf{D} is 100.0\%, which is higher than \textbf{A} $\to$ \textbf{W} (96.7\%). It is conformed with the distribution discrepancy between \textbf{A} $\to$ \textbf{W} and \textbf{W} $\to$ \textbf{D}. Therefore, the reliability of $\mathcal{A}$-distance is demonstrated.
Overall, both accuracy and $\mathcal{A}$-distance validate the superiority of our model.

\begin{table}\small
\begin{center}
\setlength{\tabcolsep}{1.0mm}{
\begin{tabular}{  l | c c  c c |c}
\toprule
Office-31&A $\to$ W&W $\to$ D&A $\to$ D&W $\to$ A&Avg. \\
\hline
\textit{Source Only}&$68.4$&$99.3$&$68.9$&$60.7$&$74.3$\\
\hline
\textit{Ours}~($w/o~\mathcal{L}_{adv}$)&$78.9$ &$99.7$ &$87.1$&$69.3$ &$83.8$ \\
\Xhline{1.2pt}
\textit{DANN}&$82.0$ &$99.1$ &$79.7$&$67.4$ &$82.1$ \\
\hline
\textit{DANN~(Em)}&$89.8$ &$\bf{100.0}$ &$90.1$&$69.0$ &$87.2$ \\
\hline
\textit{Ours}&$\bf{96.7}$ &$\bf{100.0}$ &$\bf{91.7}$&$\bf{75.0}$&$\bf{90.9}$ \\
\bottomrule
\end{tabular}}
%\end{tabularx}
\end{center}
\setlength{\abovecaptionskip}{0pt}
%\captionsetup{justification=centering}
\caption{Ablation study on the Office-31 dataset.}
%\caption{Ablation study on the Office-31 dataset, without $(w/o)$ some loss.}
\label{tab4}
\end{table}

\begin{table}\small
\begin{center}
\setlength{\tabcolsep}{1.0mm}{
\begin{tabular}{  l | c c  c c |c}
\toprule
Office-31&A $\to$ W&W $\to$ D&A $\to$ D&W $\to$ A&Avg. \\
\hline
\textit{Source Only}&$68.4$&$99.3$&$68.9$&$60.7$&$74.3$\\
\hline
\textit{Source Only~(Em)}&$89.3$&$\bf{100.0}$&$89.2$&$69.0$&$86.9$\\
\hline
\textit{DANN~(Em}$+\mathcal{L}_{tri-s}$)&$92.4$ &$99.7$ &$90.3$&$71.9$ &$88.6$ \\
\hline
\textit{DANN~(Em}$+{\mathcal{L}_{tri}}$)&$93.8$ &$99.8$ &$\bf{91.7}$&$72.6$ &$89.5$ \\
\hline
\textit{Ours}&$\bf{96.7}$ &$\bf{100.0}$ &$\bf{91.7}$&$\bf{75.0}$&$\bf{90.9}$ \\
\bottomrule
\end{tabular}}
%\end{tabularx}
\end{center}
\setlength{\abovecaptionskip}{0pt}
%\captionsetup{justification=centering}
\caption{Ablation study of focal-triplet loss.}
%\caption{Ablation study on the Office-31 dataset, without (w$/$o) some loss.}
\label{tab6}
\end{table}

\subsection{Convergence}
\begin{figure}[t]
\begin{center}
  \includegraphics[width=1\linewidth]{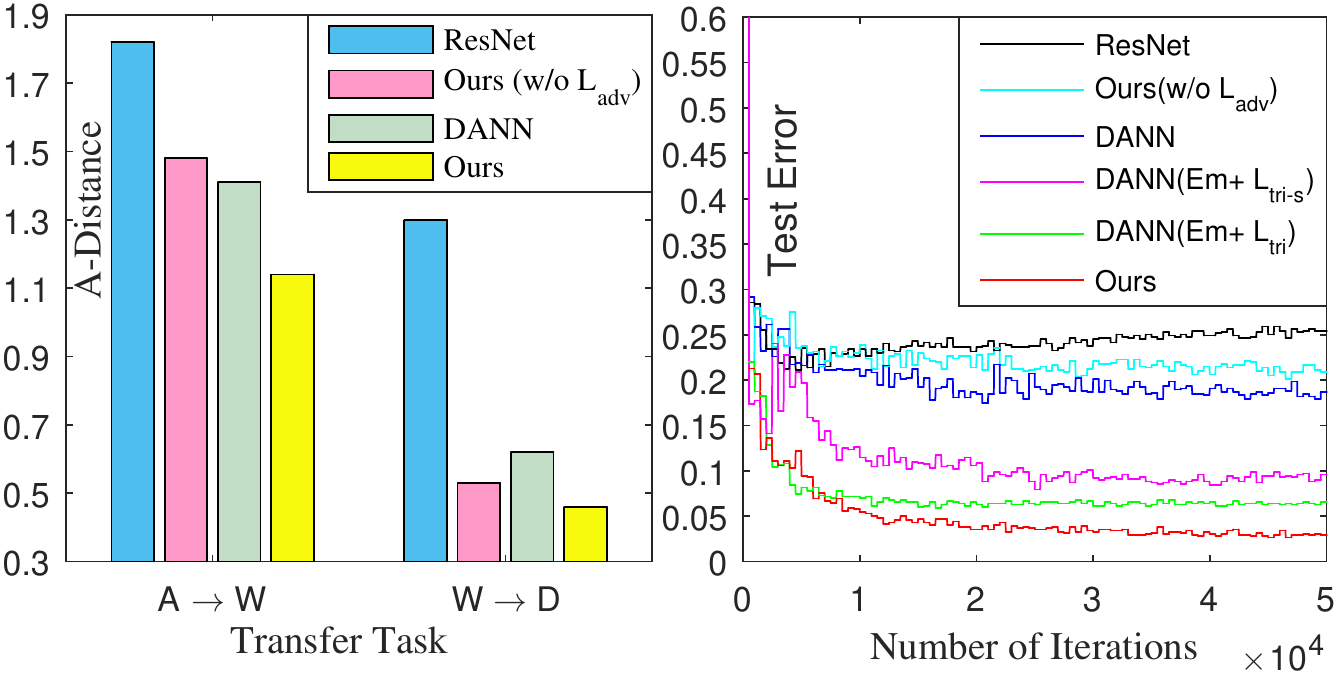}
\end{center}
%\captionsetup{justification=centering}
\setlength{\abovecaptionskip}{0pt}
   \caption{Illustration of model analysis: (a) Quantitative distribution discrepancy measured by $\mathcal{A}$-distance after domain adaptation. (b) Convergence on the test errors of different models.}
   \label{fig5}
\end{figure}

%\textbf{Convergence.}
In this section, we show the convergence of \textit{ResNet}, \textit{DANN}, \textit{Ours}~($w/o~\mathcal{L}_{adv}$), our baseline method with only source labels, \textit{i.e.}, \textit{DANN~(Em}$+\mathcal{L}_{tri-s}$), our baseline method with standard triplet, \textit{i.e.}, \textit{DANN~(Em}$+{\mathcal{L}_{tri}}$) and our complete model, respectively. The task \textbf{A} $\to$ \textbf{W} in Office-31 dataset is chosen as an example and we show the test errors (misclassification rate) of different methods with the increasing number of iterations in Figure \ref{fig5} (b). We can observe that the proposed model has a lower test error than baselines, and finally the test error of our model decreases to $0.033$ (\textit{i.e.}, 96.7\% accuracy).
Additionally, the proposed model has a comparable convergence speed with other methods, which demonstrates the competitive model complexity with baseline models.

\begin{figure*}[t]
\begin{center}
  \includegraphics[width=0.8\linewidth]{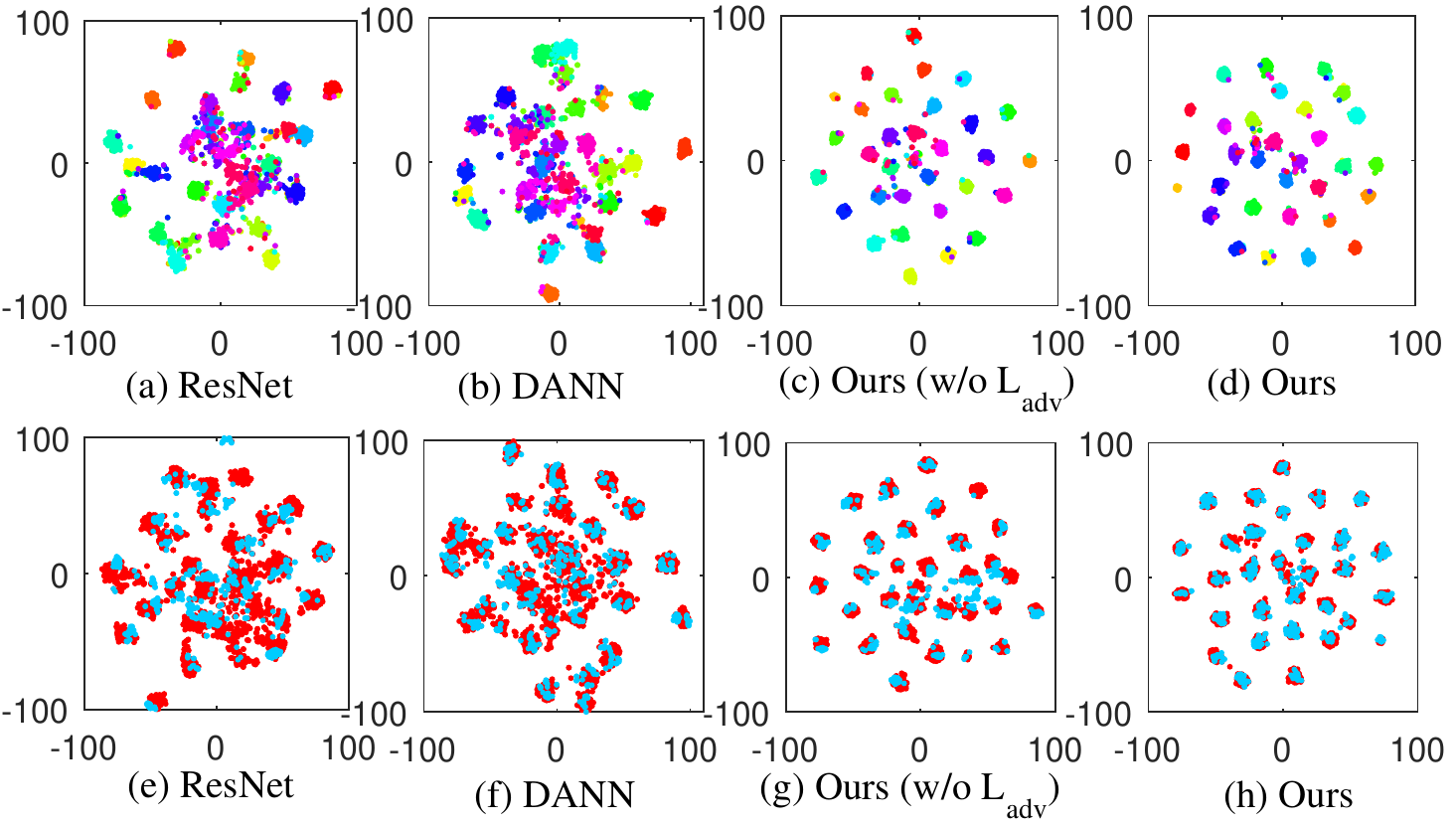}
\end{center}
  \setlength{\abovecaptionskip}{0pt}
%\captionsetup{justification=centering}
   \caption{ Feature visualization with t-SNE algorithm. \textbf{First Row}: Visualization of  \textit{Amazon} source domain feature learned by (a) ResNet, (b) DANN, (c) \textit{Ours}~($w/o~\mathcal{L}_{adv}$)~(\textit{i.e.}, without adversarial learning) and (d) Ours, respectively. \textbf{Second Row}: Visualization of source domain \textit{Amazon} ({\color{red}red}) and target domain \textit{Webcam} ({\color{blue}blue}) learned by (e) ResNet, (f) DANN, (g) \textit{Ours}~($w/o~\mathcal{L}_{adv}$) and (h) Ours (\textit{i.e.}, complete model), respectively.
   }
   \label{fig7}
\end{figure*}

\subsection{Parameter Sensitivity Analyses}
%\textbf{Parameter Sensitivity Analysis.}
We will discuss parameter sensitivity in this section. There are three items in Eq.~(\ref{eequa5})
and we leverages hyper-parameters $\lambda_1$ and $\lambda_2$ to balance the losses. Noteworthily, adversarial loss is not our focus in this paper. So we adopt the strategy to fix  $\lambda_1=1$  and change $\lambda_2$ in the range of $\{0, 0.1, 1, 2, 10\}$ to further investigate the properties of the proposed BP-triplet loss.
We conduct parameter analysis on the Office-31 and imageCLEF-DA, and the performance with respect to different trade-off parameter $\lambda_2$ is explored in Table \ref{tab5}. We can observe that the accuracy progressively grows. At most time, the best performance is achieved when $\lambda_2=1$. To avoid over-adjusting the parameters, we select the $\lambda_1=1$ and $\lambda_2=1$.
Overall, our model is generalizable to the hyper-parameters.

\subsection{Feature Visualization}
%\textbf{Feature Visualization.}

In this section, the domain invariant features learned by \textit{ResNet}, \textit{DANN}, \textit{Ours}~($w/o~\mathcal{L}_{adv}$), and our complete model are visualized to illustrate the effectiveness of our proposed model. For feature visualization, we employ the t-SNE visualization method \cite{maaten2008visualizing} on the source domain and target domain in the \textbf{A} $\to$ \textbf{W} task in Office-31 dataset. The results of feature visualization for \textit{ResNet}~(traditional CNN), \textit{DANN}~(with adversarial learning), \textit{Ours}~($w/o~\mathcal{L}_{adv}$)~(\textit{i.e.}, our model without adversarial learning), and our complete model are illustrated in Figure \ref{fig7}.

The Figure \ref{fig7} (a)-(d) represent the results in source features from 31 classes with different colors, from which we observe that \textit{Ours}~($w/o~\mathcal{L}_{adv}$) and our model can reserve better discrimination than other two baselines.
The features of target domain are visualized in Figure \ref{fig7} (e)-(h). We can observe that the features learned by \textit{ResNet} across source and target domains without considering the feature distribution discrepancy can not be well aligned. In \textit{DANN}, by considering the domain distribution alignment, the distribution discrepancy between two domains can be improved. However, as \textit{DANN} baseline method does not take the class level distribution into account, the class discrepancy of features from \textit{DANN} is not improved. In \textit{Ours}~($w/o~\mathcal{L}_{adv}$), it can alleviate domain discrepancy to some extent by similarity learning. From the results, the features learned by our model can be well aligned between two domains, but reserve more class discrimination including intra-class compactness and inter-class separability.
This evidence also accounts for that the proposed model outperforms others on a variety of unsupervised domain adaptation tasks.

\begin{table}[t]\small
\begin{center}
\setlength{\tabcolsep}{0.6mm}{
\begin{tabular}{  l | c c  c| c c c}
\toprule
Tasks&A $\to$ W&D $\to$ A&W $\to$ A&I $\to$ P&C $\to$ P&P $\to$ C \\
\hline
$\lambda_2=0$&$82.0$&$68.2$&$67.4$&$75.0$&$74.3$ &$91.5$  \\
\hline
$\lambda_2=0.1$&$89.9$ &$70.7$ &$71.9$ &$77.3$ &$77.1$ &$92.1$  \\
\hline
$\lambda_2=1$&$\bf{96.7}$ &$73.1$ &$\bf{75.0}$ &$\bf{78.6}$&$\bf{77.8}$ &$\bf{96.1}$  \\
\hline
%$\lambda_2=1$&$90.5$ &$72.2$ &$\bf{73.2}$ &$\bf{78.3}$&$76.1$ &$93.1$  \\
%\hline
$\lambda_2=2$&$93.8$ &$\bf{73.5}$ &$73.2$ &$78.3$&$77.6$ &$94.0$  \\
\hline
$\lambda_2=10$&$60.6$ &$41.3$ &$51.0$ &$56.0$ &$26.1$ &$84.8$\\
%\hline
\bottomrule
\end{tabular}}
%\end{tabularx}
\end{center}
\setlength{\abovecaptionskip}{0pt}
%\captionsetup{justification=centering}
\caption{Parameter sensitivity of our model. Classification accuracy in different datasets. The accuracy under different parameter is shown.}
%\caption{Ablation study on the Office-31 dataset, without (w$/$o) some loss.on task $A \to W$ and $I \to P$}
\label{tab5}
\end{table}

\section{Conclusion}
In this paper, we propose a metric learning based method for unsupervised domain adaptation, named as ~\textbf{BP-Triplet Net}, and it could conduct class alignment and domain alignment together. For alleviating domain bias issue, on one hand, different from previous metric learning that only consider the class-level alignment in one domain, we leverage similarity learning to achieve class-wise alignment across domains. On the other hand, in order to handle pair-wise importance imbalance and enhance the domain confusion during class alignment, we propose the BP-triplet loss deduced from Bayesian learning perspective. Additionally, we utilize the domain classifier based adversarial loss for further reducing the discrepancy distance between two domains in feature space. Simultaneously, the high-quality pseudo target labels are progressively assigned. Empirical results demonstrate that our model can effectively relieve the domain bias problem and outperform many state of the arts in various UDA tasks. In the future, we will try to extend our proposed domain adaptation method for other challenging visual tasks, such as visual relation detection~\cite{li2021interventional,shang2019annotating,di2019multiple}.

\ifCLASSOPTIONcaptionsoff
  \newpage
\fi



%\bibliographystyle{IEEEtran}
%\bibliographystyle{named}
%\bibliography{ijcai19}
%\begin{thebibliography}{1}
%\end{thebibliography}

\bibliographystyle{ieee}
\bibliography{egbib}

\end{document}